%% file: main.tex
\documentclass[final,twocolumn]{cvpr}

\usepackage{times}
\usepackage{epsfig}
\usepackage{graphicx}
\usepackage{amsmath}
\usepackage{amssymb}
\usepackage{cuted}
\usepackage{capt-of}
\usepackage{amsthm,amsmath,amssymb}
\usepackage{mathrsfs}
\usepackage{caption}
\usepackage{subcaption}
\usepackage{bbm}
\usepackage{gensymb}
\usepackage{balance}
\usepackage{adjustbox}
\usepackage{multirow}
\usepackage{media9}
\usepackage{balance}
\usepackage[toc,page]{appendix}
\usepackage[font=small,skip=0pt]{caption}
\usepackage[utf8]{inputenc}
\usepackage[T1]{fontenc}
\usepackage{enumitem}
\usepackage[nodisplayskipstretch]{setspace}
\usepackage[breaklinks=true,colorlinks,pdftex]{hyperref}

\begin{document}

\title{InMoDeGAN: Interpretable Motion Decomposition Generative Adversarial Network for Video Generation}

\author{Yaohui Wang \hskip 3em 
Francois Bremond \hskip 2em 
Antitza Dantcheva \\
Inria, Université Côte d'Azur \\
{\tt\small \{yaohui.wang, francois.bremond, antitza.dantcheva\}@inria.fr} \\
\url{https://wyhsirius.github.io/InMoDeGAN/}
}



\twocolumn[{%
\renewcommand\twocolumn[1][]{#1}%
\maketitle
\vspace{-1.2cm}
\begin{center}
\centering
\includegraphics[width=1.0\textwidth]{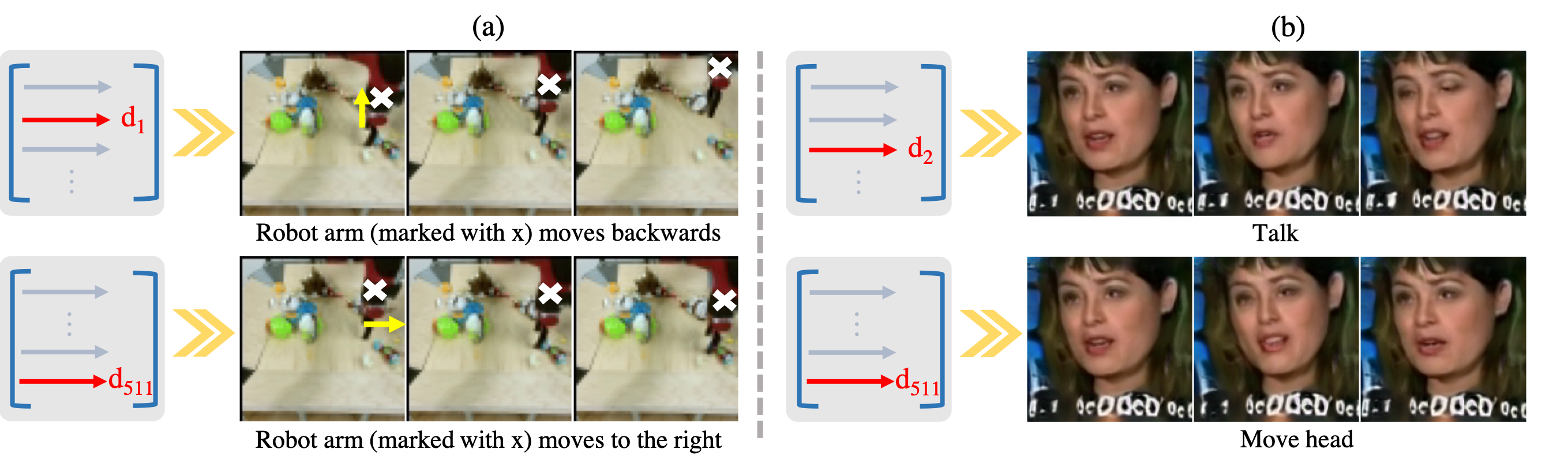}
\captionof{figure}{\textbf{Controllable video generation.} InMoDeGAN learns to decompose motion into semantic motion-components. This allows for manipulations in the latent code to invoke motion in generated videos that is human interpretable. Top (a) robot arm moves backwards, bottom (a) robot arm moves to the right. Similarly, in (b) we are animating the face to `talk' (top) and `move head' (bottom).}
\label{fig:cover}
\end{center}
}]

\begin{abstract}
In this work, we introduce an unconditional video generative model, InMoDeGAN, targeted to (a) generate high quality videos, as well as to (b) allow for interpretation of the latent space. For the latter, we place emphasis on interpreting and manipulating \textit{motion}. Towards this, we decompose motion into semantic sub-spaces, which allow for control of generated samples. We design the architecture of InMoDeGAN-generator in accordance to proposed Linear Motion Decomposition, which carries the assumption that motion can be represented by a dictionary, with related vectors forming an orthogonal basis in the latent space. Each vector in the basis represents a semantic sub-space. In addition, a Temporal Pyramid Discriminator analyzes videos at different temporal resolutions. Extensive quantitative and qualitative analysis shows that our model systematically and significantly outperforms state-of-the-art methods on the VoxCeleb2-mini and BAIR-robot datasets \wrt video quality related to  (a). Towards (b) we present experimental results, confirming that decomposed sub-spaces are interpretable and moreover, generated motion is controllable.
\end{abstract}

\section{Introduction}
Generative Adversarial Networks (GANs)~\cite{goodfellow2014generative} have witnessed remarkable progress in image generation~\cite{brock2018large,karras2017progressive,karras2019style,ledig2017photo, ma2018disentangled,miyato2018spectral,xu2018attngan,zhao2019image}. Both conditional~\cite{brock2018large,isola2017image,CycleGAN2017} and unconditional~\cite{karras2019style, Karras2019stylegan2,shaham2019singan,park2020cut} generative models have amassed exceptional capacity in generating realistic, high-quality samples. 
Most recent advances in \textit{image generation} have sought to `dissect'~\cite{bau2019gandissect} and `steer'~\cite{gansteerability} GANs by identifying a correspondence of the `inner-working' of GANs and semantic concepts in generated images. Inner-working in this context has been represented by neurons~\cite{bau2019gandissect}, as well as by latent representations~\cite{gansteerability, shen2020interpreting, voynov2020unsupervised} in \textit{pre-trained} GANs, whereas semantic concepts have included the attributes gender and age in facial image generation \cite{shen2020interpreting}, as well as camera pan and color changes in broader settings~\cite{gansteerability, goetschalckx2019ganalyze}.

Videos signify more complex data, due to the additional temporal dimension. While some research works showed early results in video generation \cite{vondrick2016generating,saito2017temporal,tulyakov2017mocogan,wang2020g3an}, related interpretability is yet to be revealed. Such interpretability and hence steerability is of particular interest, as it would render video GANs highly instrumental in a number of down-stream applications such as \textit{video editing}~\cite{wang2018vid2vid} and \textit{data augmentation}~\cite{varol17_surreal,varol19_surreact}. 
Motivated by the above, we here consider the following
question: Can we control and manipulate the complex visual world created by video GANs?

\paragraph{Contributions}
In order to answer this new and intricate question, we propose a new interpretable motion decomposing GAN for video generation, which we refer to as InMoDeGAN. In particular, we aim to interpret the latent space of InMoDeGAN by finding sub-spaces, which are endowed with semantic meanings. Once such sub-spaces have been identified, manipulating the sub-spaces allows for targeted modification of generated videos. Specifically, we here place emphasis on interpreting and modifying \textit{motion}. We note that the posed research question deviates from current efforts on interpreting \textit{appearance}~\cite{gansteerability, shen2020interpreting, voynov2020unsupervised} in the latent space.

This new problem necessitates an original architecture, streamlined to (a) generate high-quality videos, as only then an analysis of interpretability is meaningful, as well as to (b) allow for analysis of the latent motion representation. Hence, we propose a new interpretable architecture that we design based on the assumption that motion can be decomposed into independent \textit{semantic} motion-components. Therefore, we define the motion space by a linear combination of \textit{semantic} motion-components which can reflect `talking' and `robot arm moving to the right'. We implement named decomposition via a motion bank in our generator.
Once trained, InMoDeGAN allows for the incorporation/elimination of corresponding motion-components in the generated videos by activating/deactivating associated latent directions, see Fig.~\ref{fig:cover}. 

Towards (a) generating highly realistic videos, we design a two-stream discriminator, which incorporates an image discriminator, as well as a novel Temporal Pyramid Discriminator (TPD) that contains a number of video discriminators. The latter leverages on a set of temporal resolutions that are related to temporal speed. We show that while our proposed discriminator incorporates 2D ConvNets, it is consistently superior to 3D-discriminators. 
We evaluate proposed InMoDeGAN on two large datasets, namely VoxCeleb2-mini~\cite{Nagrani19} and BAIR-robot~\cite{ebert2017self}. In extensive qualitative and quantitative evaluation, we show that InMoDeGAN systematically and significantly outperforms state-of-the-art baselines \wrt video quality. In addition, we propose an evaluation framework for motion interpretability and proceed to demonstrate that InMoDeGAN is interpretable, as well as steerable. Finally, we provide experiments, where we showcase generation of both, higher-resolution, as well as longer videos.

\begin{figure*}[htbp]
\centering
\includegraphics[width=1\textwidth]{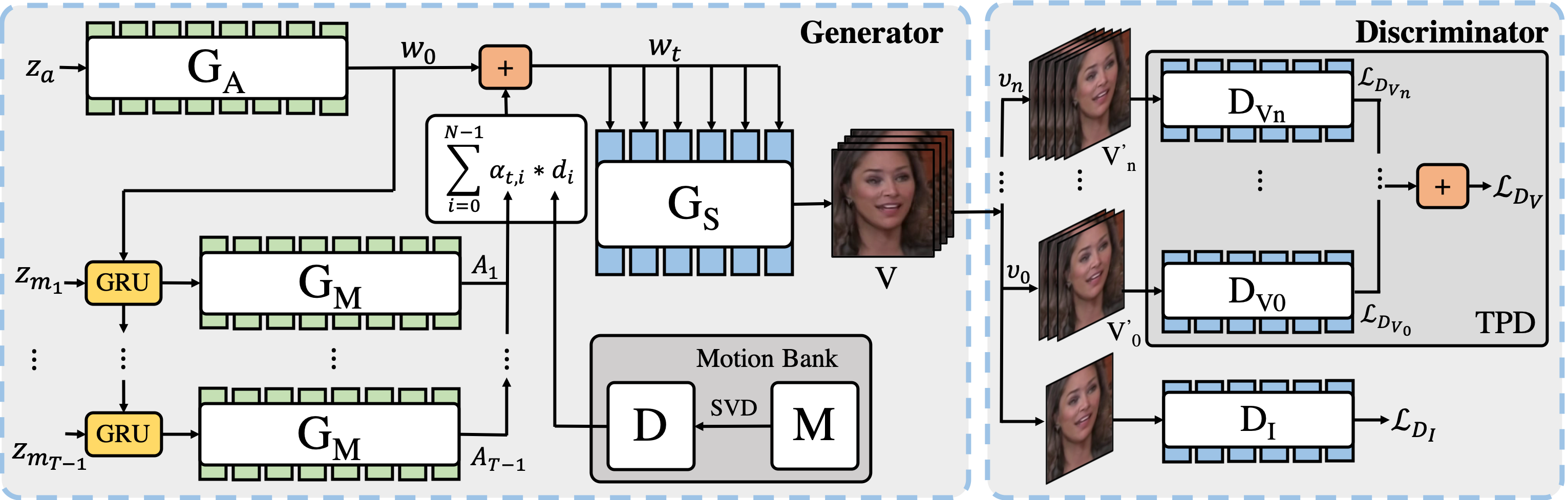}  
\caption{\textbf{InMoDeGAN-architecture.} InMoDeGAN comprises of a Generator and a two-stream Discriminator. We design the architecture of the Generator based on proposed Linear Motion Decomposition. Specifically, a motion bank is incorporated in the Generator to learn and store a motion dictionary $D$, which contains motion-directions $[d_0,d_1,..,d_{N-1}]$. We use an appearance net $G_A$ to map appearance noise $z_a$ into a latent code $w_0$, which serves as the initial latent code of a generated video. A motion net $G_M$ maps a sequence of motion noises $\{z_{m_t}\}^{T-1}_{t=1}$ into a sequence $\{A_t\}^{T-1}_{t=1}$, which represent motion magnitudes. Each latent code $w_t$ is computed based on Linear Motion Decomposition using $w_0$, $D$ and $A_t$. Generated video $V$ is obtained by a synthesis net $G_S$ that maps the sequence of latent codes $\{w_t\}^{T-1}_{t=0}$ into an image sequence $\{x_t\}^{T-1}_{t=0}$. Our discriminator comprises an image discriminator $D_I$ and a Temporal Pyramid Discriminator (TPD) that contains several video discriminators $D_{V_i}$, leveraging different temporal speeds $\upsilon_i$ to improve generated video quality. While $D_I$ accepts as input a randomly sampled image per video, each $D_{V_i}$ is accountable for one temporal resolution. }
\label{fig:architecture}
\end{figure*}

\section{Related work}
\paragraph{Image Generation.} Recent image generation methods have witnessed considerable progress~\cite{brock2018large,karras2017progressive,wang2018high,park2019semantic}. Related to our context, notably StyleGAN~\cite{karras2019style} and specifically the revised StyleGAN2~\cite{Karras2019stylegan2} constitute currently the state-of-the-art in image generation. The related architecture incorporates modulation based convolutional layers, which re-introduce a latent code at different layers of the network.
Alterations of the latent code correspond to particular manipulations in generated images. For example basic operations such as adding a vector, linear interpolation, and crossover in the latent space cause expression transfer, morphing, and style transfer in generated images. 

\paragraph{Video Generation.}  While realistic video generation is the natural sequel of image generation, it entails a number of challenges related to complexity and computation, associated to the simultaneous modeling of appearance, as well as motion. Current video generation can be categorized based on their input data into two types, \textit{unconditional} and \textit{conditional} methods.


\textit{Unconditional video generation methods} seek to map noise to video, directly and in the absence of other constraints. Examples of unconditional methods include VGAN~\cite{vondrick2016generating}, TGAN~\cite{saito2017temporal}, MoCoGAN~\cite{tulyakov2017mocogan} and G$^{3}$AN \cite{wang2020g3an}. VGAN was equipped a two-stream generator to generate foreground and background separately. TGAN firstly generated a set of latent vectors corresponding to each frame and then aimed at transforming them into actual images. MoCoGAN and G$^{3}$AN decomposed the latent representation into motion and content, aiming at controlling both factors. We note that named methods have learned to capture spatio-temporal distribution based on shallow architectures.  
Such works predominantly focused on improving the quality of generated videos, rather than exploring interpretability of the latent space.
While MoCoGAN and G$^3$AN disentangled content/appearance and motion, no further investigation on underlying semantics was provided. As opposed to that, our main goal in this paper is to gain insight into the latent space, seeking to dissect complex motion into semantic latent sub-spaces.

In contrast to unconditional video generation methods, \textit{conditional video generation methods} aim at achieving videos of high visual quality, following image-to-image generation. In this context and due to challenges in modeling of high dimensional video data, additional information such as semantic maps~\cite{pan2019video, wang2018vid2vid,wang2019fewshotvid2vid}, human key-points~\cite{jang2018video, yang2018pose, walker2017pose, chan2019everybody, zakharov2019few, wang2019fewshotvid2vid}, 3D face mesh~\cite{Zhao_2018_ECCV} and optical flow~\cite{li2018flow,ohnishi2018ftgan} have been exploited to guide appearance and motion generation. We note that given the provided motion-prior, in such methods motion cannot be further controlled.


\paragraph{GAN Interpretation.} In an effort to open the black box representing GANs, Bau \emph{et al.}~\cite{bau2019gandissect,bau2020understanding} sought to associate neurons in the generator with the encoding of pre-defined visual concepts such as colors, textures and objects. Subsequent works~\cite{shen2020interpreting, goetschalckx2019ganalyze, gansteerability, voynov2020unsupervised} proceeded to explore the interpretability of the latent space, seeking for latent representations corresponding to different semantics in generated images. \textit{Linear}~\cite{shen2020interpreting, gansteerability} and \textit{non-linear}~\cite{gansteerability} \textit{walks} in the latent space enabled for semantic concepts in the generated images to be modified. 

In this paper, we focus on unconditional video generation. Deviating from previous methods, our evolved architecture allows for high-quality video generation. We prioritize in InMoDeGAN the ability to interpret, control and manipulate \textit{motion} in generated videos. We do so by instilling a-priori the generator with a motion representation module, which learns interpretable motion-components during training, rather than interpreting a-posteriori a pre-trained generator. 

\section{Method}\label{method}
Our objective is to design an unconditional video generative model, which allows for interpretation of the latent space. While we firstly disentangle the latent space into appearance and motion, crucially, we hasten to interpret and modify the motion space. To do so, we decompose motion into semantic sub-spaces, which allow for control of generated samples.

Towards this objective, we propose in the generator $G$ a \textit{Motion bank} (see Fig.~\ref{fig:architecture}), targeted to (a) generate high quality videos, as well as (b) learn and store semantic components. The architecture of $G$ is based on proposed \textit{Linear Motion Decomposition}, which carries the assumption that motion can be represented by a dictionary with vectors forming an orthogonal basis. Each vector in the basis represents one semantic component. In addition, we propose a Temporal Pyramid Discriminator (TPD) which contains several video discriminators $D_{V_{i}}$, aiming to learn spatio-temporal distribution from different temporal resolutions. 

\subsection{Linear Motion Decomposition}\label{sec:LMD}
We formulate unconditional video generation as learning a function $G_S$ that maps a sequence of latent codes $S=\{w_t\}^{T-1}_{t=0}, w_t\sim\mathcal{W}\subset\mathbb{R}^{N}~\forall t$ to a sequence of images $V=\{x_t\}^{T-1}_{t=0}, x_{t}\sim\mathcal{X}\subset\mathbb{R}^{H\times W\times{3}}$, such that $G_S(w_t)=x_t, \forall t\in [0,T-1]$, where $T$ denotes the length of the video. $S$ is obtained by mapping a sequence of noises $Z=\{z_t\}^{T-1}_{t=0},z_{t}\sim\mathcal{Z}\subset\mathbb{R}^{N}$ into the $\mathcal{W}$ space. However, such mapping jointly learns appearance and motion, rendering $\mathcal{W}$ challenging to be interpreted. With respect to an interpretable $\mathcal{W}$, and in hindsight to our core objective, we propose to decompose motion into linear independent components. 

Given a video of high visual quality and spatio-temporal consistency, we assume that motion between consecutive frames follows a \textit{transformation} $\mathcal{T}_{t\rightarrow (t+1)}$, so that $G_{S}(w_{t+1})=\mathcal{T}_{t\rightarrow{t+1}}(G_{S}(w_{t}))$. Based on the idea of equivariance~\cite{lenc2015understanding,cohen2019gauge,hinton2011transforming}, an alteration in the latent space causes a corresponding alteration in the output, consequently a transition $\tau_{t\rightarrow{t+1}}$ affecting the latent space results in $G_{S}(\tau_{t\rightarrow{t+1}}(w_{t}))=\mathcal{T}_{t\rightarrow{t+1}}(G_{S}(w_{t}))$. 

Recent works~\cite{gansteerability, shen2020interpreting} showed that for a given image-transformation $\mathcal{T}$ such as shifting and zooming, there exists a vector $d$ in the latent space, which represents the direction of $\mathcal{T}$. By linearly navigating in this direction with a magnitude $\alpha$, a corresponding transformation $\mathcal{T}(G(w)) = G(w + \alpha * d)$ is witnessed in generated images. 

Therefore, we assume that any transition $\tau_{t\rightarrow{t+1}}$ associated to $\mathcal{T}_{t\rightarrow{t+1}}$ can be represented as a composition of motion-directions in a \textbf{motion dictionary} $D=[d_0,d_1,..,d_{N-1}], d_{i}\in \mathbb{R}^{N}$. We constrain these motion directions to form an orthogonal basis, so that 
\begin{equation}
\begin{split}
<d_i,d_j>=\left\{\begin{matrix}
 0&i \neq j\\ 
 1&i = j.
\end{matrix}\right.
\end{split}
\end{equation}
If these directions are interpretable, manipulating the magnitude of any direction should inflict a specific semantic change in the output, without affecting other directions. Therefore, in transformation
$\mathcal{T}_{{t}\rightarrow{t+1}}$, the magnitude 
$A_{t}=[\alpha_{t,0},\alpha_{t,1},...,\alpha_{t,{N-1}}],\alpha_{t,i}\in \mathbb{R}$ will vary. Each $a_{t,i}$ denotes the magnitude pertained to the $i^{th}$ direction at time step $t$. Based on this, we define the \textit{linear motion decomposition} as following
\begin{equation}\label{eq:0}
\tau_{t\rightarrow{t+1}}(w_{t})=w_{t} + \sum_{i=0}^{N-1} \alpha_{{t},i}~d_i,
\end{equation}
where the transformation between consecutive frames is indicated as
\begin{equation}\label{eq:1}
\begin{split}
G_{S}(w_{t+1})&=\mathcal{T}_{t\rightarrow{t+1}}(G_{S}(w_{t})) \\
&=G_{S}(\tau_{t\rightarrow{t+1}}(w_{t}))\\
&=G_{S}\left(w_{t} + \sum_{i=0}^{N-1} \alpha_{t,i}~d_i\right).
\end{split}
\end{equation}
The general term of $w_t$ is hence
\begin{equation}\label{eq:2}
\begin{split}
w_t = w_{0} + \sum_{i=0}^{N-1}\sum_{j=0}^{t-1}\alpha_{j,i}~d_i,~t\in[1,T-1].
\end{split}
\end{equation}
So far, we have succeeded transferring learning $w_t$ from an unknown motion space into learning three variables from three sub-spaces which contain clear meanings, namely initial appearance code $w_0$, magnitude sequence $\{A_t\}^{T-1}_{t=1}$, as well as associated motion-directions $[d_0,d_1...d_{N-1}]$. We proceed to elaborate on how we implement described linear motion decomposition in our architecture. 

\subsection{Generator}\label{sec:gen}
The initial latent code $w_0$ serves as a representation of \textit{appearance} in the first and all following frames of an output video. At the same time, the vector $A_t$ represents a set of magnitudes associated to motion directions in a transition and hence is accountable for \textit{motion}. Taking that into account, we decompose $\mathcal{Z}$ into two separated spaces $\mathcal{Z_A}$ and $\mathcal{Z_M}$, which represent appearance and motion, respectively. Hence $w_0$ is generated by mapping an appearance noise $z_a\sim \mathcal{Z_A}$ 
using an appearance net $G_A$. $A_t$ is mapped from the motion noise $z_{m_{t}}\sim \mathcal{Z_M}$ by a motion net $G_M$. In order to ensure temporal consistency in the latent space, we integrate a GRU~\cite{cho2014learning} with its initial code set to be $z_a$ prior to the mapping. We note that $G_A$ and $G_M$ are two different 8-layer MLPs.

Based on our \textit{linear motion decomposition}, the motion dictionary $D$ is entitled to an orthogonal basis. 
We propose to find a matrix, with eigenvectors representing $d_i$. More specifically, we pre-define a matrix $M \in \mathbb{R}^{N\times N}$ and devise it trainable, updating it along with the parameters in the generator.  $D$ is represented as the transpose of \textit{right singular vectors} of $M$, $M = U\Sigma V^{T}$ and $D=V^{T}$. Each $d_i$ is an eigenvector of matrix $M^{T}M$ and is learned based on adversarial learning. Once trained, $M$ captures the motion distribution of the training dataset and decomposes it into $N$ independent directions. We show that some directions are interpretable and moreover can be manipulated, which results in related modifications of generated results, see Sec.~\ref{exp:motion_bank}. $M$ is initialized randomly and updated with other parameters in $G$ via back-propagation. We refer to $M$ and $D$ jointly as \textbf{motion bank}.

We adapt the architecture proposed by Karras \etal~\cite{Karras2019stylegan2} in $G_S$. We note that $G_S$ serves as a rendering network, which incorporates a sequence of convolutional blocks aiming to up-sample a learned constant into high resolution images. In each block, convolutional layers are modulated by the respective input $w_t$, in order to learn different appearances. Each $w_t$ is computed according to Eq.~\ref{eq:2} and serves as input to $G_S$ to generate related frame $x_t = G_S(w_t)$.

\subsection{Discriminator}\label{sec:dis}
Temporal speed in videos has been a pertinent cue in action recognition~\cite{feichtenhofer2019slowfast, yang2020temporal}. We note that videos sampled at temporal speeds $\upsilon$, which represent temporal resolutions, provide a set of motion features. For this reason, we propose a Temporal Pyramid Discriminator (TPD) that leverages videos of different temporal resolutions in order to ensure high video quality in generation. 

Principally, our discriminator follows the two-stream architecture of MoCoGAN~\cite{tulyakov2017mocogan} and G$^3$AN~\cite{wang2020g3an}. We have a stream comprising an image discriminator $D_I$, as well as a stream incorporating the proposed TPD. While the input of $D_I$ is a randomly sampled frame, TPD accepts as input a full video sequence. TPD includes a number of video discriminators $D_{V_i}$, each $D_{V_i}$ is accountable for one temporal resolution.

Deviating from previous work~\cite{tulyakov2017mocogan, wang2020g3an}, we here propose to leverage 2D ConvNets in $D_V$ rather than 3D ConvNets. We apply time to channel (TtoC) to concatenate sampled frames in channel dimension, in order to construct a video sampled at speed $\upsilon_i$ into an image $V^{'}_i\in \mathbb{R}^{H\times W\times K}$, where $\frac{K}{3}$ denotes the number of sampled frames. We surprisingly find that such design can substantially improve the visual quality, while ensuring temporal consistency of generated videos. We report experimental results in Sec.~\ref{sec:exp}.

\subsection{Learning}
We use non-saturating loss~\cite{goodfellow2014generative} with $\mathcal{R}_{1}$ regularization~\cite{Mescheder2018ICML,Karras2019stylegan2} as our objective function following the setting of StyleGAN2~\cite{Karras2019stylegan2}. The loss of TPD combines the losses of each video discriminator $D_{V_i}$ in the pyramid,  $\sum_{i=0}^{n-1}\mathcal{L}_{D_{V_i}}$. We optimize the network based on the full objective
\begin{equation}
\min_{G}\left(\lambda\sum_{i=0}^{n-1}\max_{D_{V_i}}\mathcal{L}_{D_{V_i}} + \max_{D_I}\mathcal{L}_{D_{I}}\right),
\end{equation}
where $n$ is a hyperparameter denoting the number of video discriminators to be used during training. We empirically identify appropriate $n$ values in our two datasets, see Sec.~\ref{sec:exp}. $\lambda$ aims to balance the loss between $D_I$ and TPD.

\section{Experiments and Analysis}\label{lab:experiments}
We present extensive experiments, which include the following. In \textit{video quality evaluation}, we quantitatively evaluate the ability of InMoDeGAN to generate realistic videos and compare related results with four state-of-the-art methods for unconditional video generation. We then analyze the effectiveness of the proposed TPD. In addition, we provide an ablation study, which indicates the appropriate number of temporal resolutions for both datasets. 

In \textit{interpretability evaluation}, we aim to discover interpretable directions in the motion dictionary. Towards this, we propose a new evaluation framework that quantifies motion in generated videos based on optical flow. We show that directions in the motion dictionary, based on our proposed framework, are indeed semantically meaningful. Further, we demonstrate that generated videos can be easily modified by manipulating such directions. Notably, our model allows for controllable video generation based on pre-defined trajectories for different directions.  

Finally, we conduct further analysis of high resolution generation, linear interpolation and go beyond training data to explore longer video generation. 

\paragraph{Implementation details.} We implement InMoDeGAN using PyTorch~\cite{paszke2019pytorch}. All experiments are conducted on 8 V100 GPUs (32GB) with total batch size 32 (4 videos per GPU). We use Adam optimizer~\cite{kingma2014adam} with a learning rate 0.002 and set $\beta_1$ = 0.0, $\beta_2$ = 0.99. Dimensions of $z_a$ and $z_m$ are set to be 512 and 256, respectively. We pre-define to learn $N=512$ directions in the motion dictionary, the dimension of each direction is set to be 512. $\lambda$ is set to be 0.5 for all experiments. In TPD, we use four time steps 1,3,5,7 to sample videos on VoxCeleb2-mini and three time steps 1,3,5 on BAIR-robot. More implementation and training details are described in Sec.~\ref{app:architecture}.

\subsection{Datasets and evaluation metric}
we report evaluation results on following two datasets.

\textbf{VoxCeleb2-mini.} We construct a subset of VoxCeleb2~\cite{Nagrani19}, which comprises of over 1 million videos pertaining to 6000 celebrities, talking in different real-world scenarios containing diverse complex motions (\eg, head moving, talking, camera zooming, etc.). As the original dataset includes redundant scenarios, we construct a new subset of 12000 videos, where we randomly select video sequences pertaining to 20 diverse videos per each of the 6000 subjects. We note that videos include large appearance diversity.

\textbf{BAIR-robot~\cite{ebert2017self}.} The dataset incorporates a single-class and depicts stationary videos of a robot arm moving and pushing a set of objects. We use the training set of this dataset which contains 40000 short videos.

\textbf{Evaluation metric.} We use video FID~\cite{NIPS2017_7240} to quantitatively evaluate visual quality and temporal consistency in generated videos. For the computation, we appropriate ResNeXt-101~\cite{hara2018can} pre-trained on Kinetics-400~\cite{Carreira_2017_CVPR} as feature extractor and take features before last fully connected layer to compute the FID. We randomly sample 10000 videos to compute the values for each experiment.

\subsection{Video quality evaluation}\label{sec:exp}
We firstly compare InMoDeGAN with four state-of-the-art methods, namely VGAN, TGAN, MoCoGAN, as well as G$^3$AN. We generate videos pertained to named methods with spatial resolution of $64\times64$ and temporal length of 32 for VGAN and 16 for the other methods. 
Related FIDs are reported in Tab.~\ref{tab:fid_sota}. InMoDeGAN systematically outperforms other methods \wrt video quality by obtaining the lowest FID on both datasets. This is a pertinent prerequisite for latent space interpretation, as only highly realistic videos would allow for a meaningful interpretation. We show generated samples on our project \href{https://wyhsirius.github.io/InMoDeGAN/}{website}.

\begin{table}[thb]
\centering
\setlength{\tabcolsep}{3.5pt}
\setlength\arrayrulewidth{1pt}
\begin{tabular}{ccccc}
\hline
Method & {\textbf{VoxCeleb2-mini}} & {\textbf{BAIR-robot}} \\
\hline
VGAN~\cite{vondrick2016generating} & 38.13 & 147.23 \\
TGAN~\cite{saito2017temporal} & 23.05 & 120.22 \\
MoCoGAN~\cite{tulyakov2017mocogan} & 12.69 & 13.68 \\
G$^{3}$AN~\cite{wang2020g3an} & 3.32 & 1.58 \\
\hline
InMoDeGAN & \textbf{2.37} & \textbf{1.31} \\
\hline
\end{tabular}
\caption{\textbf{Comparison of InMoDeGAN with four state-of-the-art models.} InMoDeGAN systematically and significantly outperforms other methods on both datasets \wrt FID. The lower FID, the better video quality.}\label{tab:fid_sota}
\end{table}

\begin{figure*}[htbp]
\centering
\begin{subfigure}{0.49\textwidth}
\centering
\includegraphics[width=0.49\textwidth]{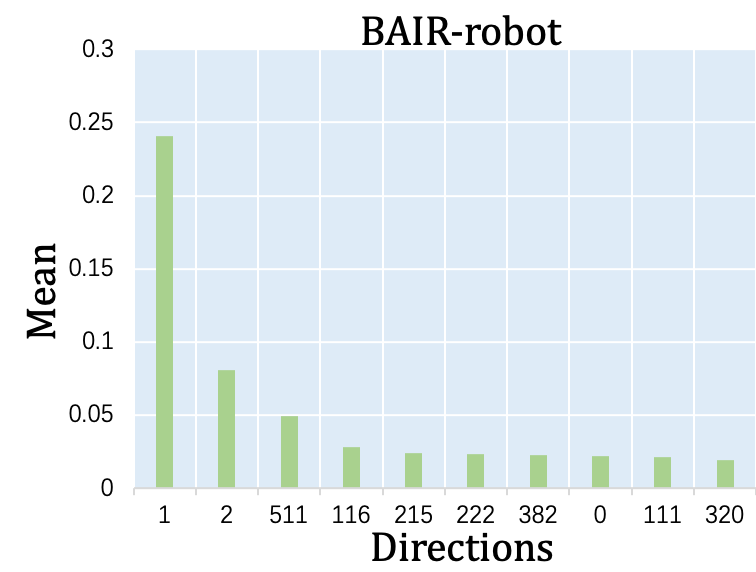}
\includegraphics[width=0.49\textwidth]{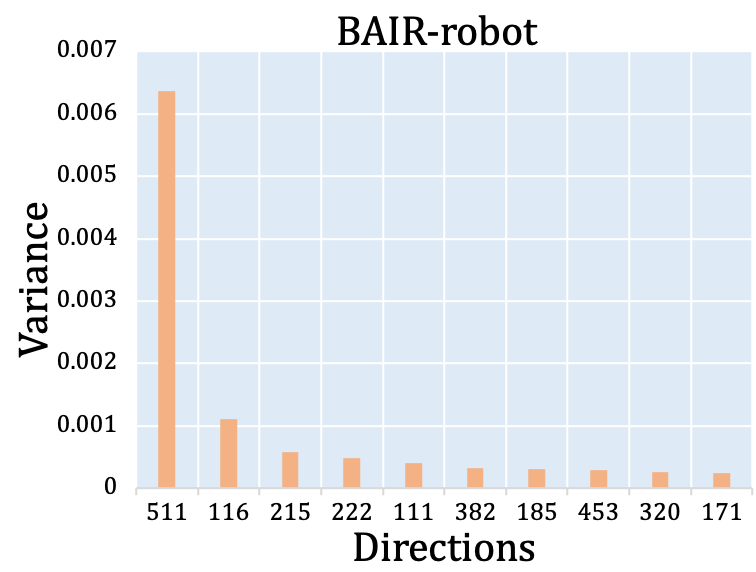}
\includegraphics[width=0.49\textwidth]{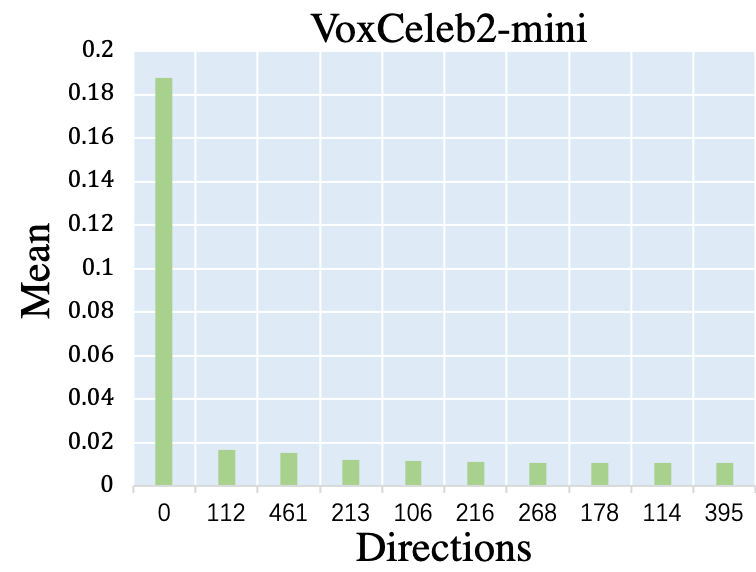}
\includegraphics[width=0.49\textwidth]{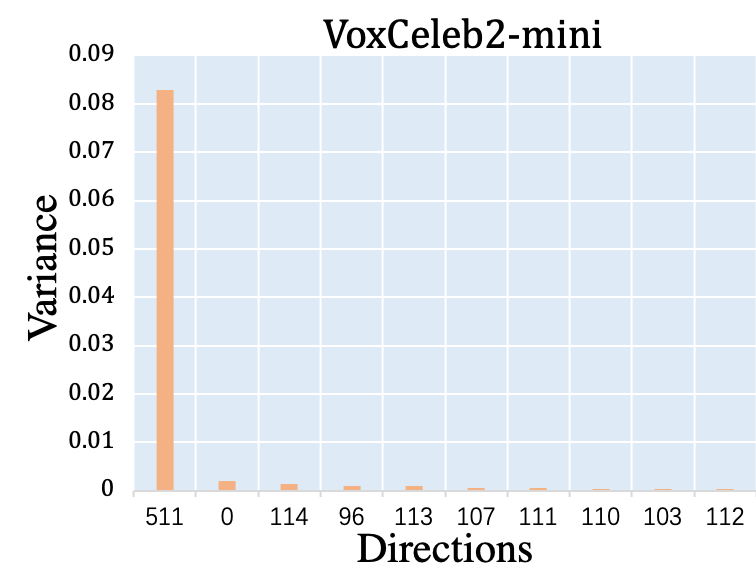}
\caption{\textbf{Mean and variance of $A_{\Bar{t}}$.}}
\label{subfig:variance_mean}
\end{subfigure}
\begin{subfigure}{0.49\textwidth}
\centering
\includegraphics[width=0.49\textwidth]{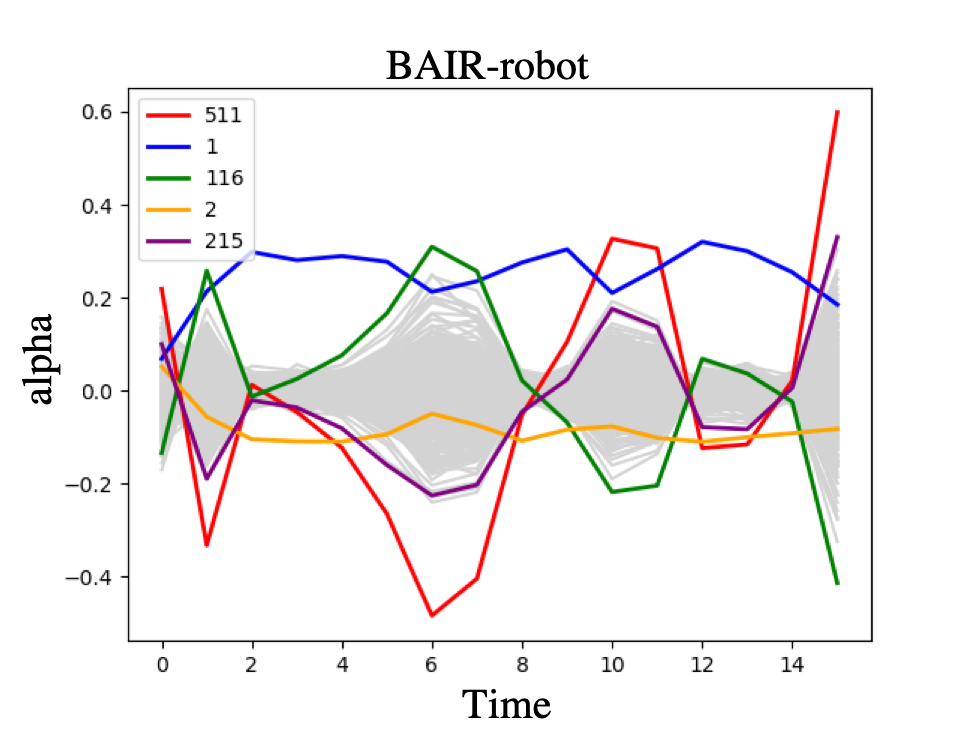}
\includegraphics[width=0.49\textwidth]{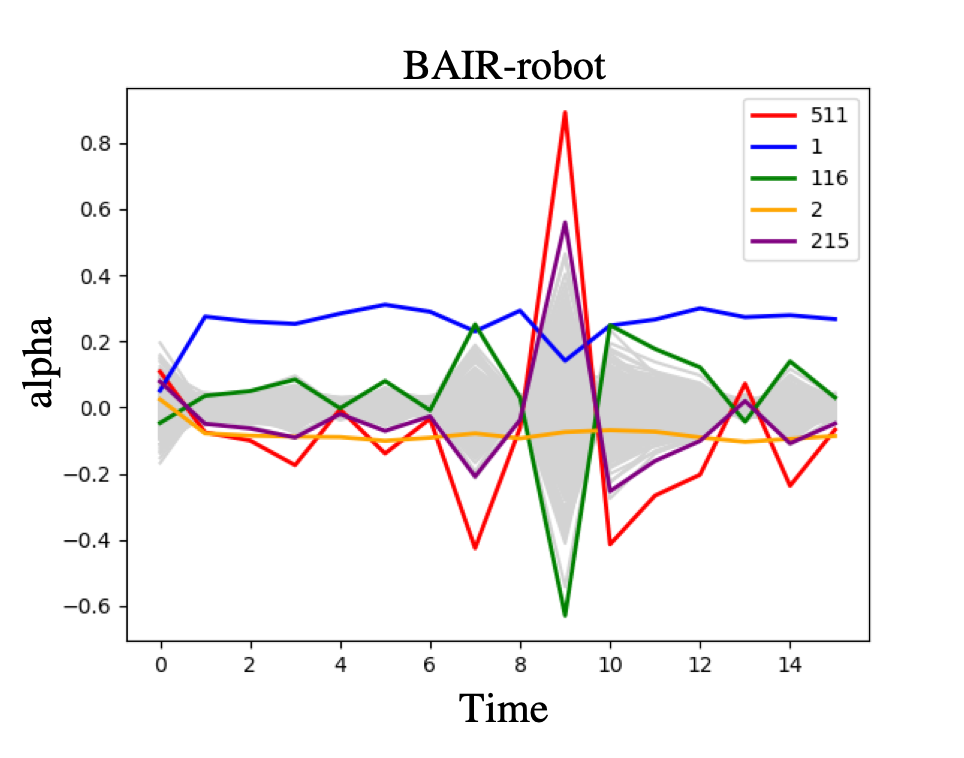}
\centering
\includegraphics[width=0.49\textwidth]{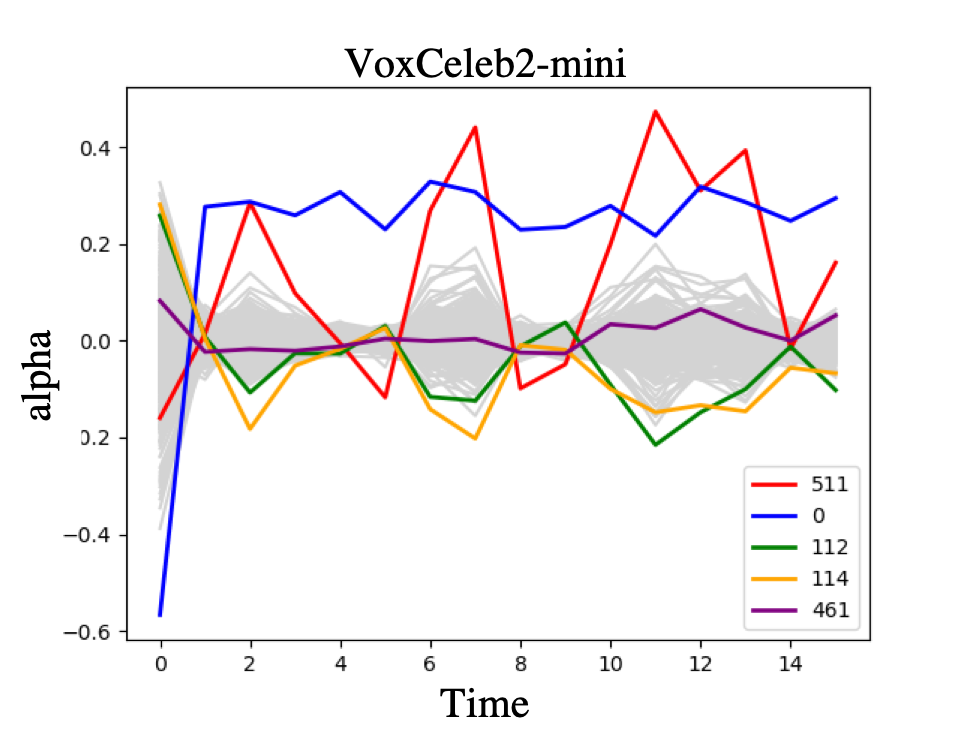}
\includegraphics[width=0.49\textwidth]{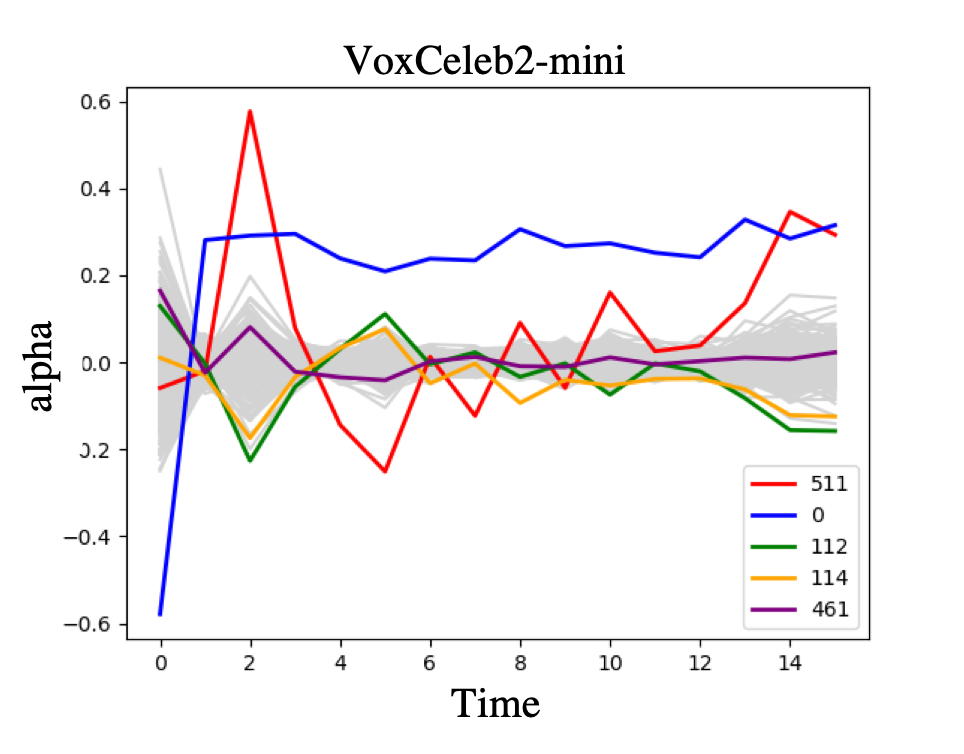}
\caption{\textbf{Time v.s. $\alpha$}}
\label{subfig:alpha}
\end{subfigure}
\caption{\textbf{Analysis of $\alpha$.} (a) Mean and variance bar charts, indicating top 10 motion-directions with highest values in $A_{\Bar{t}}$. (b) Time v.s. $\alpha$. Each figure represents a video sample. We illustrate two samples from BAIR-robot (top) and two from VoxCeleb2-mini (bottom). Top 5 dimensions in $\alpha$ are plotted in different color.}
\label{fig:stat}
\end{figure*}

\textbf{Effectiveness of TPD.} We replace the original 3D discriminators in VGAN, TGAN, MoCoGAN, as well as G$^3$AN with TPD, maintaining all training configurations as in the previous experiment. We report FIDs related to original and proposed discriminators in all algorithms and both datasets in Tab.~\ref{tab:fid_dis_eva}. We observe that TPD improves the results of all methods significantly and consistently. This confirms that videos sampled with a set of temporal resolutions contain different features, which are beneficial in the discriminator. 

On a different but related note, we observe during training that models without image discriminator (VGAN and TGAN) tend to reach mode collapse rapidly on BAIR-robot (high FID in Tab.~\ref{tab:fid_dis_eva}). This is rather surprising, as BAIR-robot constitutes the simpler of the two datasets, comprising videos of a robot arm moving, with a fixed background. The occurrence of very similar scenes might be the reason for the challenging distinguishing of real and fake spatial information in the absence of an image discriminator.

\begin{table}[htb]
\setlength{\tabcolsep}{3.2pt}
\setlength\arrayrulewidth{1pt}
\centering
\begin{tabular}
{cccccc}
\hline
\multirow{2}{*}{Method} &\multicolumn{2}{c}{\textbf{VoxCeleb2-mini}} &
\multicolumn{2}{c}{\textbf{BAIR-robot}} \\ \cline{2-5}
& 3D & TPD & 3D & TPD \\
\hline
VGAN~\cite{vondrick2016generating} & 38.13 & 16.33 & 147.23 & 93.71 \\
TGAN~\cite{saito2017temporal} & 23.05 & 21.24 & 120.22 & 120.04 \\
MoCoGAN~\cite{tulyakov2017mocogan} & 12.69 & 7.07 & 13.68 & 3.16 \\
G$^3$AN~\cite{wang2020g3an} & 3.32 & 2.98  & 1.58 & 1.50 \\
\hline
\end{tabular}
\caption{\textbf{Evaluation of TPD.} When replacing the initial 3D discriminator with TPD, the latter significantly and consistently improves the FID of all 4 state-of-art models for the VoxCeleb2-mini and BAIR-robot datasets.}
\label{tab:fid_dis_eva}
\end{table}

In addition, we conduct an \textbf{ablation study}, seeking to determine the optimal number of temporal resolutions in TPD for both datasets. Associated results are reported in Tab.~\ref{tab:fid_dis_type}, which suggest that while for VoxCeleb2-mini, which contains complex motion, we achieve the lowest FID on four temporal resolutions, for BAIR-robot, which is simpler \wrt occurring motion, three resolutions suffice.

\begin{table}[htb]
\setlength{\tabcolsep}{3.2pt}
\setlength\arrayrulewidth{1pt}
\centering
\begin{tabular}
{ccc}
\hline
TPD type &\textbf{VoxCeleb2-mini} & \textbf{BAIR-robot} \\
\hline
$D_{V_{0}}$, $D_{V_{1}}$, $D_{V_{2}}$, $D_{V_{3}}$ & \textbf{2.37} & 1.56 \\
$D_{V_{0}}$, $D_{V_{1}}$, $D_{V_{2}}$, & 2.65 & \textbf{1.31} \\
$D_{V_{0}}$, $D_{V_{1}}$ & 2.76 & 1.33 \\
$D_{V_{0}}$ & 2.84 & 1.58 \\
\hline
\end{tabular}
\caption{\textbf{Ablation study on video discriminators in TPD.} Number of video discriminators associated to temporal resolutions. FID is reported for comparison. Lower FID indicates a superior quality of generated videos.}
\label{tab:fid_dis_type}
\end{table}

\begin{figure*}[htbp]
\includegraphics[width=1.0\textwidth]{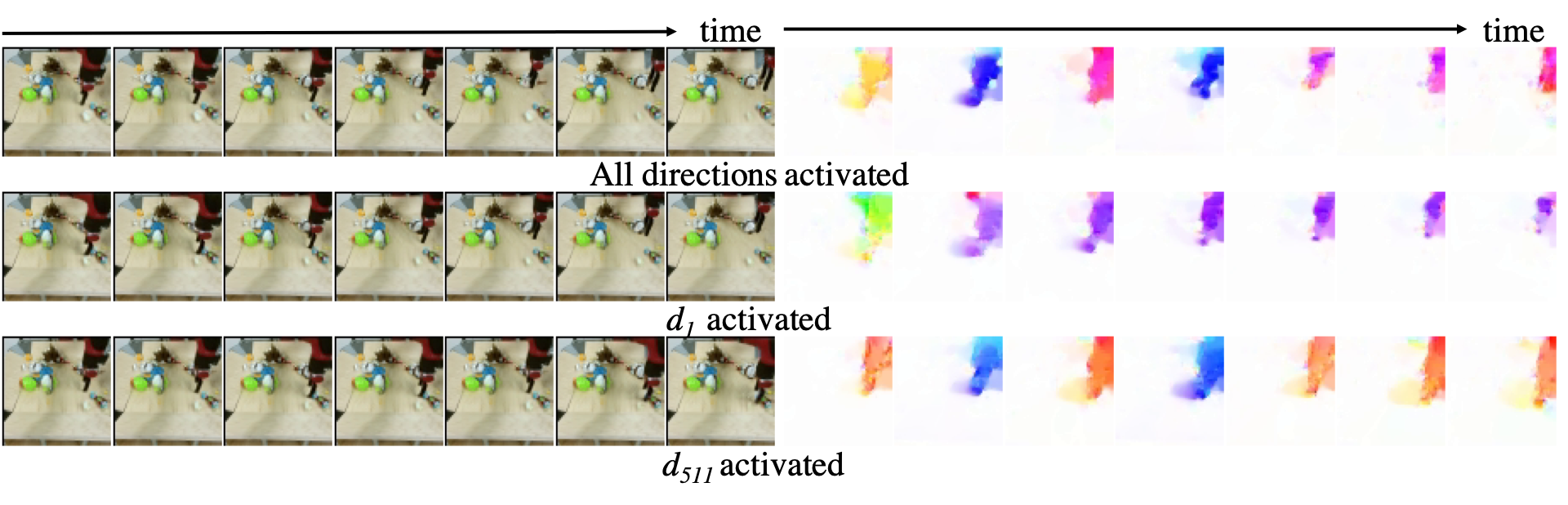}
\caption{\textbf{Directions analysis on BAIR-robot.} A generated video sample, related optical flow images (top), activation of \textit{only} $d_1$ (middle), and activation of \textit{only} $d_{511}$ (bottom). Optical flow images indicate that $d_1$ is accountable for moving the robot arm backward, whereas $d_{511}$ for moving it left and right.}
\label{fig:bair_motion_flow}
\end{figure*}
\begin{figure}[htbp]
\centering
\begin{subfigure}{0.11\textwidth}
\centering
\includegraphics[width=1.0\textwidth]{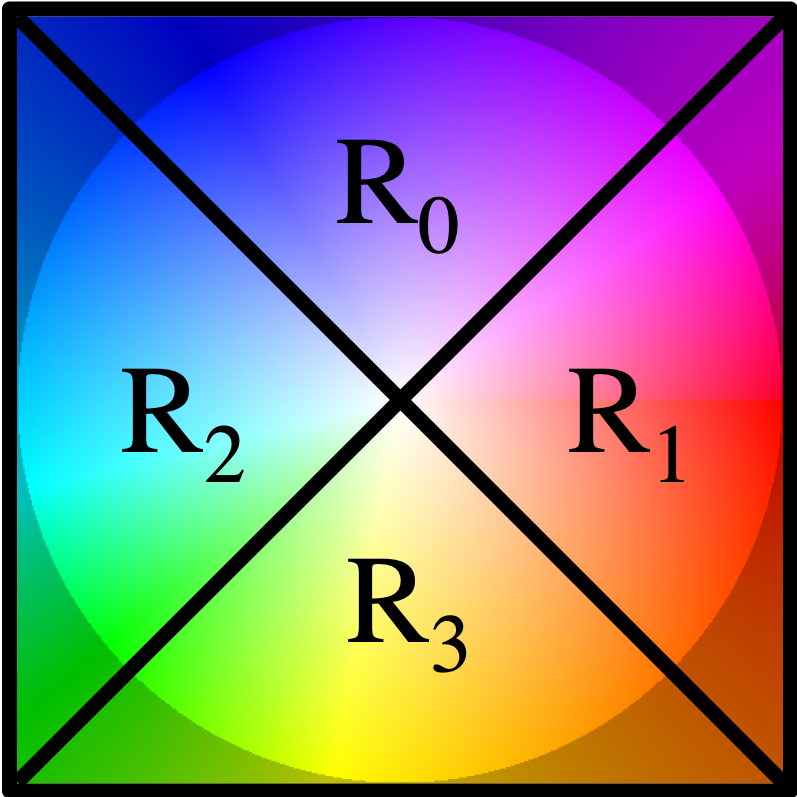}
\caption{}
\end{subfigure}
\begin{subfigure}{0.11\textwidth}
\centering
\includegraphics[width=1.0\textwidth]{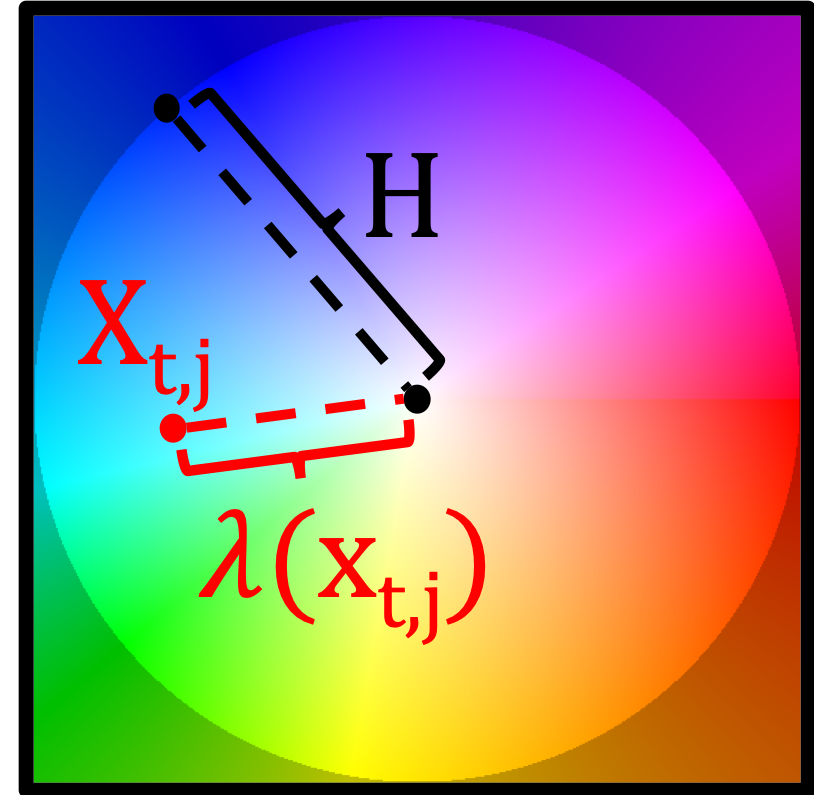}
\caption{}
\end{subfigure}
\begin{subfigure}{0.11\textwidth}
\centering
\includegraphics[width=1.0\textwidth]{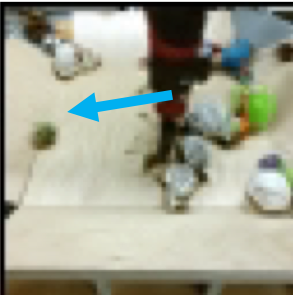}
\caption{}
\end{subfigure}
\begin{subfigure}{0.11\textwidth}
\centering
\includegraphics[width=1.0\textwidth]{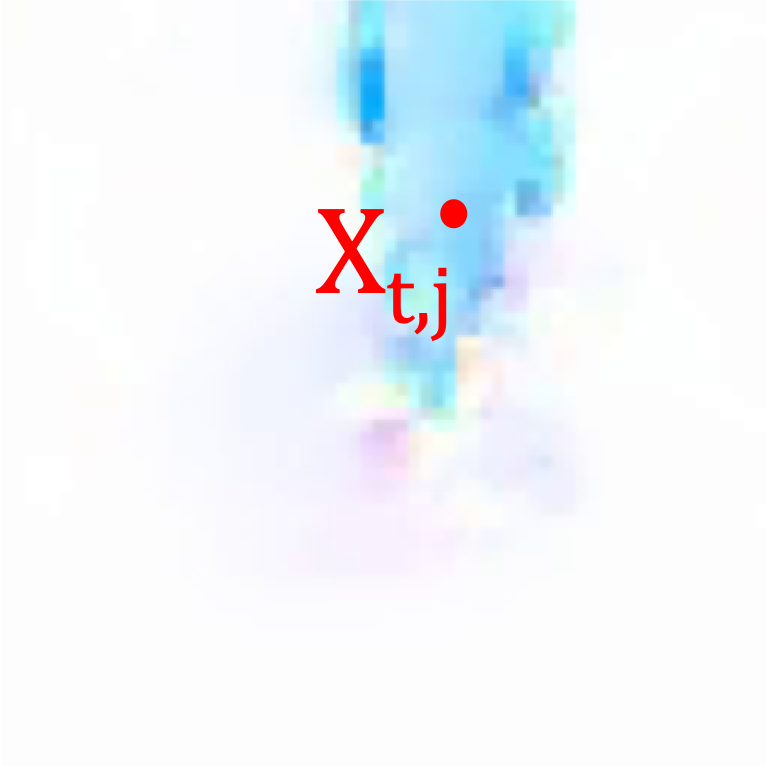}
\caption{}
\end{subfigure}
\caption{\textbf{Optical flow quantization.} (a) Middlebury colorwheel, (b) $\lambda(x_{t,j})$ and H on the colorwheel, (c) one frame from BAIR-robot and (d) related optical flow.}
\label{fig:color_wheel}
\end{figure}

\subsection{Interpretability evaluation}\label{exp:motion_bank}
Above, we have provided experimental proof that InMoDeGAN is able to generate high quality videos. In this section, we focus on discussing, how to leverage those videos to find interpretable directions in the motion dictionary. Towards this, firstly we analyze $\alpha$, seeking to find directions with highest impact.

Then, we present our proposed evaluation framework for quantifying motion, in order to find semantic meaning of such directions. Next, we show generated results based on manipulation of such directions. Finally, we demonstrate that our model allows for controllable generation by navigating in found interpretable directions in pre-defined trajectories. 

\paragraph{Do all directions contribute equally?} As per Eq.~\ref{eq:2}, each $\alpha_{j,i}$ indicates the magnitude of $d_i$ at time step $j$. We sample $10000$ latent codes as evaluation set and compute mean and variance over time, for the full set, in order to obtain
$A_{\Bar{t}}=[\alpha_{\Bar{t},0},\alpha_{\Bar{t},1},...,\alpha_{\Bar{t},{N-1}}],\alpha_{\Bar{t},i}\in \mathbb{R}$.
Fig.~\ref{subfig:variance_mean} shows mean and variance values of the 10 most pertinent dimensions in $A_{\Bar{t}}$ for both datasets. We note that for both datasets, $\alpha_{\Bar{t},511}$ has the largest variance, which indicates that $d_{511}$ leads to the strongest motion variation in generated videos. At the same time, $\alpha_{\Bar{t},1}$ (BAIR-robot) and $\alpha_{\Bar{t},0}$ (VoxCeleb2-mini) encompass highest mean values, respectively. Therefore, we have that $d_{1}$ (BAIR-robot) and $d_{0}$ (VoxCeleb2-mini) show high and continuous magnitudes, respectively.

Moreover, we are interested in the course of each $\alpha_{j,i}$ over time, which we portray in Fig.~\ref{subfig:alpha}. Specifically, we randomly select two samples per dataset and highlight a set of $\alpha_{0:15,i}$ in different colors. We have that, while $\alpha_{0:15,511}$ (in red) has the largest amplitude in both datasets, $\alpha_{0:15,1}$ (BAIR-robot) and $\alpha_{0:15,0}$ (VoxCeleb2-mini) (in blue) maintain high but steady values over time, respectively. This supports our findings, as displayed in Fig.~\ref{subfig:variance_mean}. 

Based on the above, we conclude that directions in the motion dictionary \textit{do not} contribute equally in composing motion.

\begin{figure*}[htbp]
\includegraphics[width=1.0\textwidth]{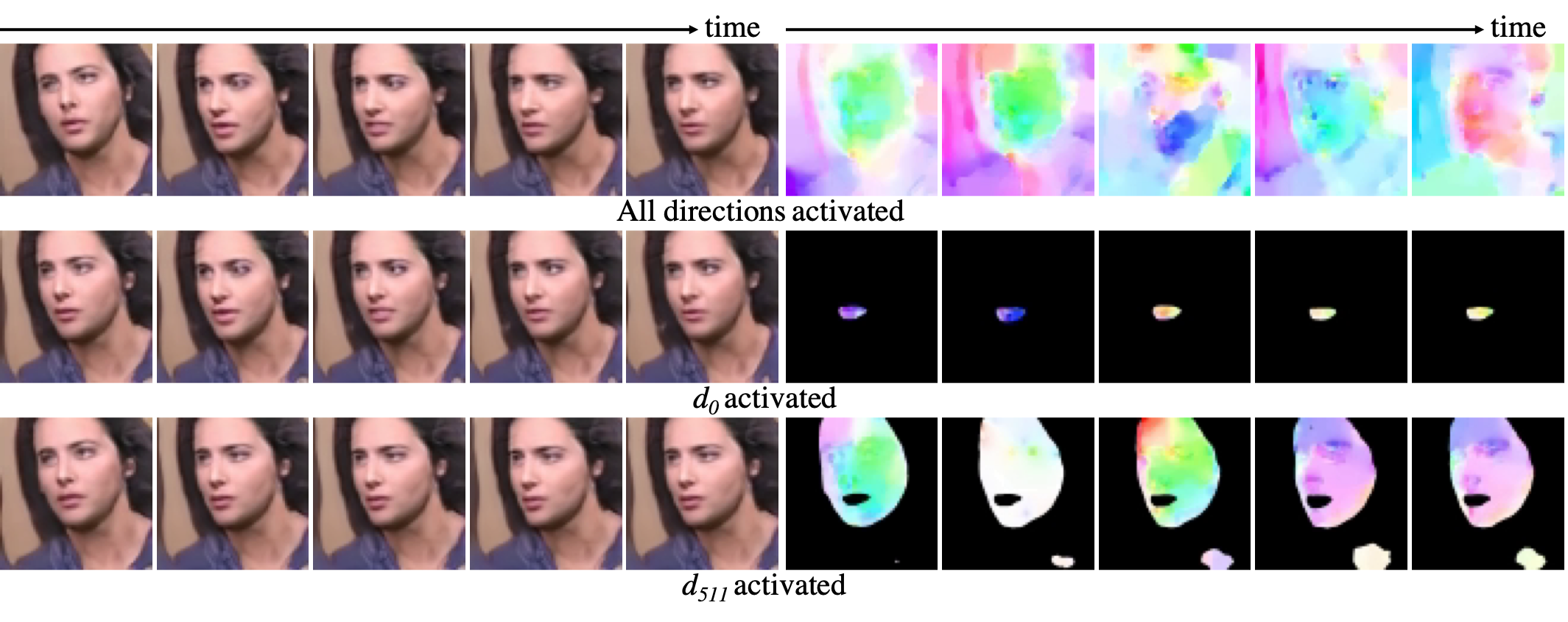}
\caption{\textbf{Direction analysis in VoxCeleb2-mini.} A generated video sample and associated optical flow images (top), by \textit{only} activating $d_0$ (middle), and by \textit{only} activating $d_{511}$ (bottom). While $d_0$ controls the mouth region, $d_{511}$ controls the head region.}
\label{fig:vox_motion_flow}
\end{figure*}
\begin{figure}[htbp]
\includegraphics[width=0.47\textwidth]{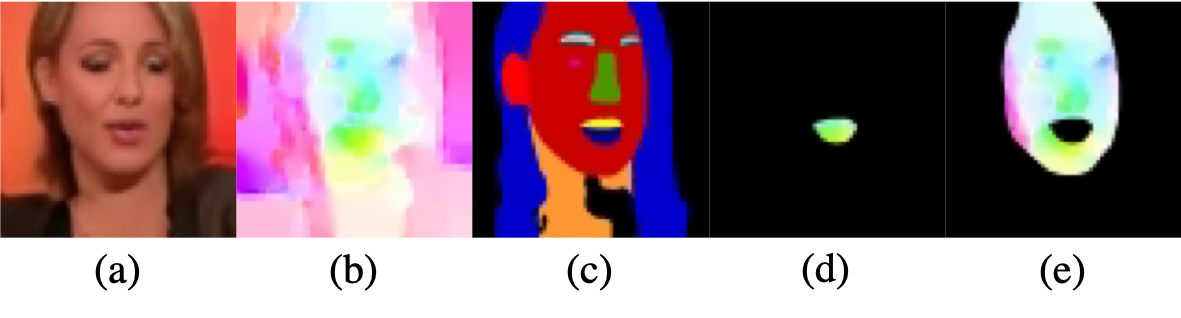}
\caption{\textbf{Global and local motion extraction.} (a) Generated image, (b) related optical flow, (c) semantic map, (d) mouth-flow image, and (e) face-flow image based on training with VoxCeleb2-mini.}
\label{fig:img_mask}
\end{figure}

\paragraph{Are motion components interpretable?}
We here aim to semantically quantify motion directions by a novel framework using optical flow. Firstly, we represent the optical flow according to the Middlebury evaluation \cite{baker2011database}. Specifically, we partition the flow into four histogram bins, namely $R_0$, $R_1$, $R_2$ and $R_3$, to cover the $360\degree$ range of orientation and amplitude, see Fig.~\ref{fig:color_wheel}. While different motion directions are represented by the hue values, motion magnitude is indicated by the brightness. Hence each $R_i$ represents a motion range. Next, for any given optical flow video, we quantify motion in $R_i$ as following.
\begin{equation}\label{eq:motion_quant_1}
\begin{split}
\phi_{i}&=\frac{1}{N_i}\sum_{t=0}^{T-1}\sum_{j=0}^{N-1} \frac{\lambda{(x_{t,j})}}{H}\mathbbm{1}_{R_{i}}(x_{t,j}), i\in \{0,1,2,3\}, \\
\end{split}    
\end{equation}
with total motion in the video being computed as
\begin{equation}\label{eq:motion_quant_2}
\begin{split}
\Phi&=\frac{1}{N}\sum_{i=0}^{3}\sum_{t=0}^{T-1}\sum_{j=0}^{N-1} \frac{\lambda{(x_{t,j})}}{H}\mathbbm{1}_{R_{i}}(x_{t,j}), \\
\end{split}    
\end{equation}
where $x_{t,j}$ denotes the value of the $j^{th}$ pixel at time step $t$ in an optical flow video, which contains $N$ color pixels in total. $N_i$ denotes the total number of color pixels in $R_i$. $\lambda(x_{t,j})$ measures the distance from $x_{t,j}$ to the center of the colorwheel, whose radius is $H$ (see Fig.~\ref{fig:color_wheel}). A large $\phi_i$ indicates a frequent and strong appearance of motion associated to $R_i$.

For BAIR-robot, we proceed to evaluate the set of directions $d_1$, $d_2$, $d_{116}$ and $d_{511}$, as they exhibit the highest impact according to Fig.~\ref{subfig:variance_mean}. Our idea is to quantify the motion difference $\Delta{\phi_{i}}={{\phi^{d_k}_{i}-\phi_{i}}}$ in each $R_i$, when $d_k$ is deactivated (set $\alpha_k=0$ ) in original videos. 

We sample 1000 videos and deactivate each of the chosen directions, respectively, building an evaluation dataset containing 6000 samples (1000 original + 5000 deactivated). We report averaged $\phi_i$ over the full evaluation set for each region in Tab.~\ref{tab:motion_quant}. When $d_1$ is deactivated, motion in $R_0$ and $R_3$ are strongly reduced. Similarly for $d_{511}$,  $\phi_1$ and $\phi_2$ obtain the largest decline. We note that for some directions motion changes are minor. As $(R_0,R_3)$ and $(R_1,R_2)$ are opposite regions, $d_1$ and $d_{511}$ represent symmetric motions. 
To illustrate this, we generate samples by \textit{only} activating $d_1$ and \textit{only} activating $d_{511}$, respectively, while maintaining other directions deactivated. Fig.~\ref{fig:bair_motion_flow} shows one sample and related optical flow, from which we deduce that the results match our quantitative evaluation, which suggested that $d_1$ represents `robot arm moving back and forth', and $d_{511}$ represents `robot arm moving left and right'.

\begin{table}[htb]
\setlength{\tabcolsep}{3.2pt}
\setlength\arrayrulewidth{1pt}
\centering
\begin{tabular}
{ccccc}
\hline
& $\Delta{\phi_0}$ & $\Delta{\phi_1}$ & $\Delta{\phi_2}$ & $\Delta{\phi_3}$ \\
\hline
$d_1$ & \textbf{-0.008} & 0.017 & 0.002 & \textbf{-0.033} \\
$d_2$ & -0.001 & 0.002 & 0.002 & -0.005 \\
$d_{116}$ & 0.000 & -0.001 & 0.001 & 0.000 \\
$d_{511}$ & 0.007 & \textbf{-0.087} & \textbf{-0.059} & 0.019 \\
\hline
\end{tabular}
\caption{\textbf{$\Delta{\phi_i}$ on BAIR-robot.} Motion difference in four regions ($R_0$, $R_1$, $R_2$, $R_3$) caused by deactivating motion-directions. }
\label{tab:motion_quant}
\end{table}

VoxCeleb2-mini comprises a more complex dataset than BAIR-robot. Related videos contain concurrent global motion (\eg head moving, camera zooming), as well as local motion (talking). For VoxCeleb2-mini we therefore analyze global and local motion by focusing specifically on head and mouth regions, computing facial semantic maps, and further head-flow and mouth-flow videos for each sample (see Fig.~\ref{fig:img_mask}). We use the method of Yu\etal~\cite{yu2018bisenet} to extract facial semantic maps.

For VoxCeleb2-mini we proceed to select the top 4 directions $d_0$, $d_{112}$, $d_{114}$, and $d_{511}$ from Fig.~\ref{subfig:variance_mean} and sample 1000 videos for evaluation. Deviating from above, we here quantify video motion changes in head ${\Delta\Phi_{head}}$ and mouth regions $\Delta\Phi_{mouth}$, respectively. Tab.~\ref{tab:motion_quant_vox} shows that while deactivation of $d_{511}$ triggers the largest motion decline in the head region, the deactivation of $d_0$ leads to the largest decline of mouth-motion. Considering that head movement contributes to mouth movement, we compute $\Delta{\Phi_{mouth}}-\Delta{\Phi_{head}}$, excluding global from local motion. However, $d_0$ still remains highest contribution to mouth motion. Similar to BAIR-robot, we illustrate samples by activating \textit{only} $d_0$, and \textit{only} $d_{511}$, respectively, in Fig.~\ref{fig:vox_motion_flow}. While $d_0$ reflects mouth motion, $d_{511}$ represents head motion. This is conform to our quantitative evaluation.

Therefore, we verify that some directions in our motion dictionary are interpretable. In addition, we are able to control motion by (de-)activating such directions.  

\begin{table}[htb]
\setlength{\tabcolsep}{3.2pt}
\setlength\arrayrulewidth{1pt}
\centering
\begin{tabular}
{cccc}
\hline
& $\Delta{\Phi_{head}}$ & $\Delta{\Phi_{mouth}}$ & $\Delta{\Phi_{mouth}}$-$\Delta{\Phi_{head}}$ \\
\hline
$d_0$ & -0.012 & \textbf{-0.052} & \textbf{-0.040} \\
$d_{112}$ & -0.001 & -0.005 & -0.005 \\
$d_{114}$ & -0.000 & -0.005 & -0.005 \\
$d_{511}$ & \textbf{-0.036} & -0.027 & 0.008 \\
\hline
\end{tabular}
\caption{\textbf{$\Delta{\Phi_{head}}$ and $\Delta{\Phi_{mouth}}$ on VoxCeleb2-mini.} Motion difference in head and mouth regions induced by deactivation of motion-directions.}
\label{tab:motion_quant_vox}
\end{table} 

As we have already found interpretable directions, we show for BAIR-robot, by providing pre-defined trajectories to $d_1$ and $d_{511}$, that we are able to generate videos in a controllable manner. We provide detailed experimental description in Sec.~\ref{app:further} and show results generated results on project website.

\subsection{Further analysis}
We here experiment with \textbf{linear interpolation} in the latent space, see Sec.~\ref{app:results}. We note that such interpolations are evidence that InMoDeGAN has learned a smooth mapping from the latent space to real videos, rather than memorized training data.

Moreover, we show that our model generalizes well to \textbf{high-resolution video generation}. Towards this, we generate $128\times 128$ videos, as trained on VoxCeleb2-mini, as well as on \textbf{UCF101}~\cite{soomro2012ucf101}. In this context, we repeat the interpretability evaluation and observe again interpretable directions related to mouth and head motion. For UCF101, we conduct quantitative evaluation based on a metric proposed by TGANv2~\cite{TGAN2020}. We report evaluation results of VoxCeleb2-mini ($128\times 128$) in Tab.~\ref{tab:vox128-fid} and UCF101 in Tab.~\ref{tab:ucf}. Results show that our method outperforms current state-of-the-art on UCF101 by exhibiting lower FID and higher IS.

Finally, we generate \textbf{longer videos} to explore the limit of our model for the VoxCeleb2-mini and BAIR-robot datasets. InMoDeGAN is able to generate videos of frame-length beyond training data (16 frames), reaching up to around 32 frames on VoxCeleb2-mini and 45 frames on BAIR-robot, which are highly promising. Generated results are shown on project website and experimental details are described in Sec.~\ref{app:results}.

\section{Conclusions}
We have presented a novel video generative model, InMoDeGAN, which is aimed at (a) generating high quality videos, as well as (b) allowing for interpretation of the latent space.
In extensive evaluation on two datasets, InMoDeGAN outperforms quantitatively and qualitatively state-of-the-art methods \wrt visual quality. Crucially, we have shown the ability of InMoDeGAN to decompose motion in semantic sub-spaces, enabling direct manipulation of the motion space. We have showcased that proposed Temporal Pyramid Discriminator, streamlined to analyze videos at different temporal resolutions, while involving only 2D ConvNets, outperforms 3D counterparts. In further analysis we have explored generation of longer videos, as well as of videos with higher resolution. Future work involves the analysis of our method on more complex human activity datasets, where we intend to investigate the possibility to control motion of each joint in a human body.

{\small
\bibliographystyle{ieee_fullname}
\balance
\bibliography{egbib}
}

\include{app}

\end{document}

%% file: app.tex











\begin{appendices}
We describe related settings for generation in Sec.~\ref{app:results}. Then, we show high-resolution ($128\times 128$) videos generation results on VoxCeleb2-mini and UCF101 datasets and proceed to compare proposed InMoDeGAN with state-of-the-art unconditional GANs in Sec.~\ref{app:further}. Finally, we present additional implementation details of our proposed InMoDeGAN in Sec.~\ref{app:architecture}. 

\section{Generated results}\label{app:results}
Generated videos pertain to following generation settings are shown on our project \href{https://wyhsirius.github.io/InMoDeGAN/}{website}.

\begin{figure*}
\begin{subfigure}{0.47\textwidth}
\centering
\includegraphics[width=0.9\textwidth]{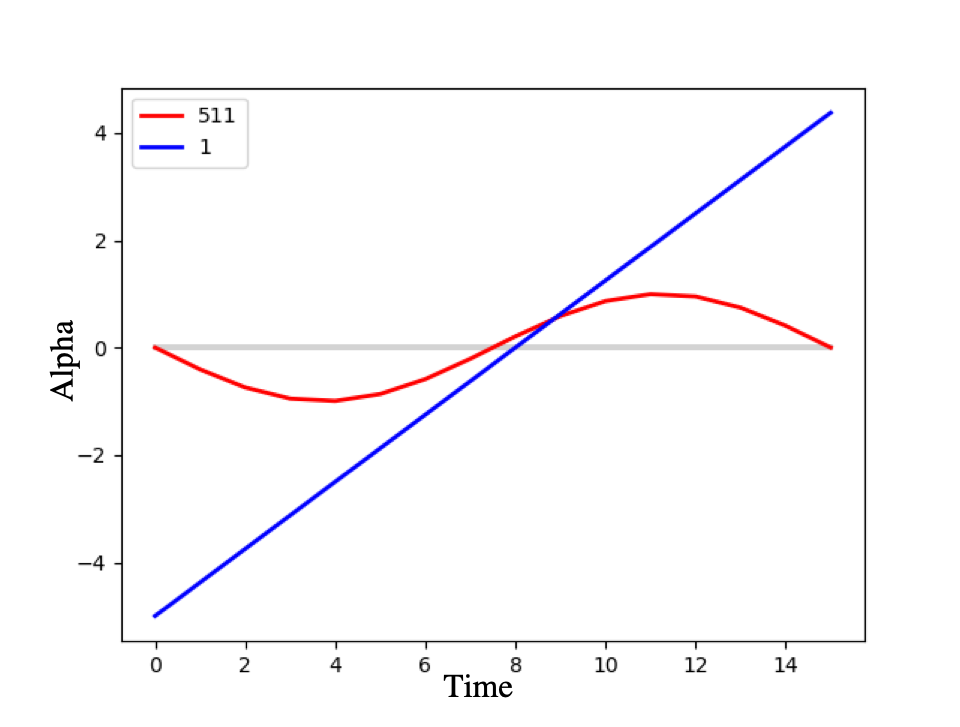}
\caption{Trajectory type 1.}
\label{subfig:trj1}
\end{subfigure}
\begin{subfigure}{0.47\textwidth}
\centering
\includegraphics[width=0.9\textwidth]{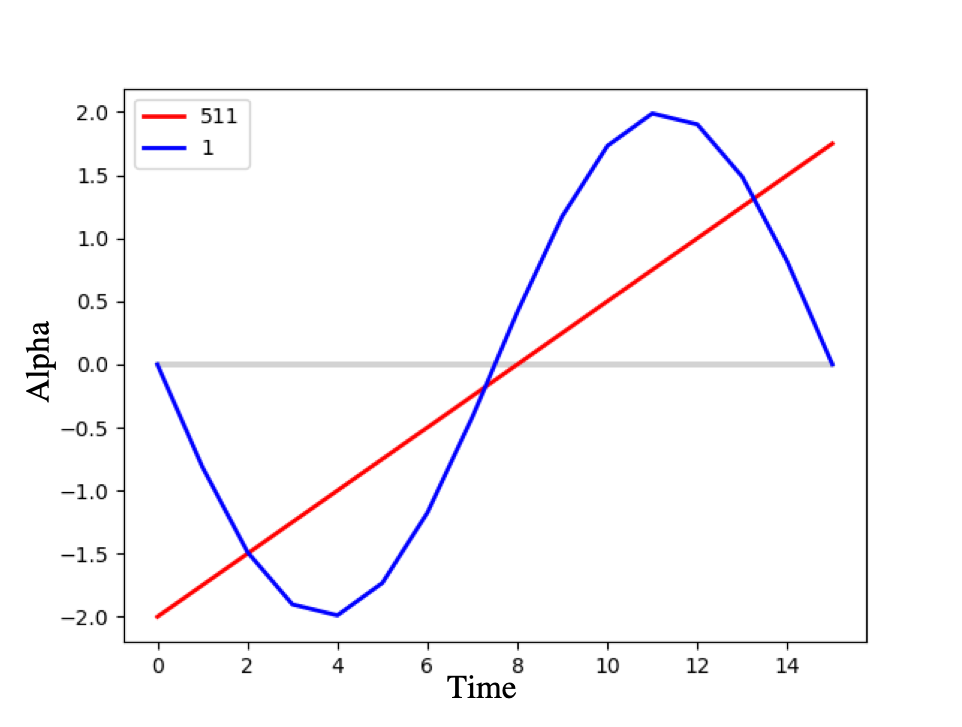}
\caption{Trajectory type 2.}
\label{subfig:trj2}
\end{subfigure}
\caption{\textbf{Two pre-defined trajectories.} (a) We provide a \textit{linear} trajectory for $d_1$ and a \textit{sinusoidal} trajectory for $d_{511}$. (b) We provide a \textit{sinusoidal} trajectory for $d_1$ and a \textit{linear} trajectory for $d_{511}$.}
\end{figure*}

\paragraph{1. Random generation.} We randomly sample different appearance noise vectors $z_a$ and motion noise sequences $\{z_{m_t}\}^{T-1}_{t=0}$ on both datasets, respectively. We present results on VoxCeleb2-mini in two resolutions ($128\times 128$ and $64\times 64$) and BAIR-robot.
\paragraph{2. Motion decomposition.} Towards demonstrating the identified interpretable directions in InMoDeGAN, we illustrate generated videos by activating these directions for both datasets. $All$ indicates generated videos are obtained by activating \textit{all} directions in the motion bank, whereas $d_{i}$ denotes that \textit{only} the $i^{th}$ direction has been activated. For VoxCeleb2-mini (\textit{$128\times 128$}), based on proposed evaluation framework, we find that $d_2$ and $d_{511}$ in the motion bank are entailed with highest magnitudes. According to presented optical flow evaluation, we determine the relation to \textit{mouth} and \textit{head} regions, respectively. We provide generated videos, in which $d_2$ and $d_{511}$ have been activated, respectively. We observe that indeed $d_2$ represents \textit{talking}, whereas $d_{511}$ corresponds to \textit{head moving}. Similarly, for VoxCeleb2-mini (\textit{$64\times 64$}), we have quantitatively proven in the main paper that motion related to mouth and head are represented by $d_0$ and $d_{511}$, respectively.  Generated videos verify this result. For BAIR-robot, we show generated videos with $d_1$ and $d_{511}$ activated, respectively. We provide corresponding optical flow videos, which illustrate the moving directions of the robot arm when the two directions are activated. We note that while $d_1$ moves the robot arm \textit{back and forth}, $d_{511}$ moves it \textit{left and right}.
\paragraph{3. Appearance and motion disentanglement.} For each dataset, we show videos generated by combining one appearance noise vector $z_a$ and 16 motion noise vectors $\{z_{m_0},z_{m_1},...,z_{m_{15}}\}$. The associated generated videos contain notable \textit{motion diversity}, as the same appearance is being animated by different motions.  
\paragraph{4. Linear interpolation.} We linearly interpolate two appearance codes, $z_{a_{0}}$ and $z_{a_{1}}$, and associate each intermediate appearance to one motion code sequence. Results show that intermediate appearances are altered gradually and smoothly. Notably, we observe continuous changes of head pose, age and cloth color in videos related to VoxCeleb2-mini; as well as changes of starting position and background in videos related to BAIR-robot.
\paragraph{5. Controllable generation.} While we train our model in unconditional manner, surprisingly we find that by providing different $\alpha$-trajectories as semantic directions, we are able to create videos with controllable motion, following such trajectories (for BAIR-robot). We show two types of $\alpha$-trajectories over time for $d_1$ and $d_{511}$ in Fig.~\ref{subfig:trj1} and Fig.~\ref{subfig:trj2}, respectively. While in Fig.~\ref{subfig:trj1} a \textit{linear} trajectory is provided for $d_{1}$ and a \textit{sinusoidal} trajectory for $d_{511}$, in Fig.~\ref{subfig:trj2},  $d_1$ and $d_{511}$ are activated oppositely. We illustrate generated videos by activating $d_1$, $d_{511}$, as well as both directions, respectively, while all other directions maintain deactivated (set $\alpha$ to 0). The related results indicate that the robot arm can indeed be controlled directly with different trajectories. 
\paragraph{6. Longer video generation.} Interestingly, despite that our model is trained with 16-frame videos, our model is able to generate results beyond the length of the training data ($>$16 frames). In order to explore the limit of our model, we generate longer videos by providing as input more than 16 vectors of motion noise sequences for both datasets. Specifically, for BAIR-robot, in each instance we input the size 16, 32 and 48 of $z_{m_i}$, in order to generate videos with different temporal lengths. We note that in generated videos of length about 45 frames the robot arm disappears. For VoxCeleb2-mini, which incorporates more complex motion, we find that after 32 frames, generated frames become blurry and ultimately faces melt.

Nevertheless, this is highly encouraging, as we only utilize GRU for temporal refinement. In future work, we plan to explore advanced global temporal modeling modules such as Transformers~\cite{vaswani2017attention}, towards replacing the GRU~\cite{cho2014learning} for longer generation. 

\section{Evaluation of high-resolution video generation}\label{app:further}
We evaluate our generated high-resolution ($128\times 128$) videos pertained to both, VoxCeleb2-mini and UCF101~\cite{soomro2012ucf101}. We use the evaluation protocol introduced in the main paper for VoxCeleb2-mini. Results are reported in Tab.~\ref{tab:vox128-fid}. Naturally, higher resolution corresponds to better (lower) FID.

Towards a fair comparison with state-of-the-art results on UCF101, we use the evaluation protocol introduced in TGANv2~\cite{TGAN2020}. It uses a C3D~\cite{tran2015learning} that has been pre-trained on UCF101 as feature extractor. We report video results \wrt Inception Score (IS) and Fr\'echet Inception Distance (FID) in Tab.~\ref{tab:ucf}. Our method outperforms other methods using both evaluation metrics \wrt high-resolution video generation.  

\begin{table}[thb]
\centering
\setlength{\tabcolsep}{3.5pt}
\setlength\arrayrulewidth{1pt}
\begin{tabular}{cccc}
\hline
Method & FID ($\downarrow$) \\
\hline
VGAN~\cite{vondrick2016generating} ($64\times 64$) & 38.13  \\
TGAN~\cite{saito2017temporal} ($64\times 64$) & 23.05 \\
MoCoGAN~\cite{tulyakov2017mocogan} ($64\times 64$) & 12.69 \\
G$^{3}$AN~\cite{Wang_2020_CVPR} ($64\times 64$) & 3.32 \\
\hline
InMoDeGAN ($64\times 64$) & 2.37  \\
InMoDeGAN ($128\times 128$) & \textbf{0.25}  \\
\hline
\end{tabular}
\caption{
{Comparison of InMoDeGAN with four state-of-the-art models.} InMoDeGAN systematically outperforms the other models on \textbf{VoxCeleb2-mini} \wrt FID. }
\label{tab:vox128-fid}
\end{table}

\begin{table}[htb]
\setlength{\tabcolsep}{3.2pt}
\setlength\arrayrulewidth{1pt}
\centering
\begin{tabular}
{ccc}
\hline
Method & IS ($\uparrow$) & FID ($\downarrow$) \\
\hline
VGAN~\cite{vondrick2016generating} & $8.31 \pm .09$ & - \\
TGAN~\cite{saito2017temporal} & $11.85 \pm .07$ & - \\
MoCoGAN~\cite{tulyakov2017mocogan} & $12.42 \pm .03$ & - \\
ProgressiveVGAN~\cite{acharya2018towards} & $13.59 \pm .07$ & - \\
TGANv2~\cite{TGAN2020} & $26.60 \pm .47$ &  $3431 \pm 19$ \\
\hline
InMoDeGAN & $\mathbf{28.25 \pm .05}$ & $\mathbf{3390 \pm 83}$ \\
\hline
\end{tabular}
\caption{{Comparison of InMoDeGAN with five state-of-the-art models.} InMoDeGAN systematically outperforms the other models on \textbf{UCF101} \wrt IS and FID. Values are taken from~\cite{TGAN2020} except InMoDeGAN.}
\label{tab:ucf}
\end{table} 

\section{Additional implementation details}\label{app:architecture}
We design a new architecture, endowed with the ability to interpret the \textit{motion space}. For appearance, we adapt the synthesis net $G_S$ from StyleGAN2~\cite{Karras2019stylegan2}. However, we find that original \textit{image-level} layer-wise \textit{noise inputs} bring flickering problem in generated results. Towards solving this, we propose \textit{video-level} layer-wise \textit{noise inputs}. We provide as input a set of noise vectors for one video, and all frames share the same noise in the same convolutional layer. We find that such modification allows for generating smoother videos in contrast to the original implementation. In the discriminator, $D_I$ maintains the same architecture as the original implementation in ~\cite{Karras2019stylegan2} for image generation. In each $D_{V_i}$, we modify the input channel dimension from 3 into $K$, where $\frac{K}{3}$ denotes the frame number for each sampled video. We have that different temporal speed results in different $K$. 

\end{appendices}



%% file: main.bbl
\begin{thebibliography}{10}\itemsep=-1pt

\bibitem{acharya2018towards}
Dinesh Acharya, Zhiwu Huang, Danda~Pani Paudel, and Luc Van~Gool.
\newblock Towards high resolution video generation with progressive growing of
  sliced wasserstein gans.
\newblock {\em arXiv preprint arXiv:1810.02419}, 2018.

\bibitem{baker2011database}
Simon Baker, Daniel Scharstein, JP Lewis, Stefan Roth, Michael~J Black, and
  Richard Szeliski.
\newblock A database and evaluation methodology for optical flow.
\newblock {\em International journal of computer vision}, 92(1):1--31, 2011.

\bibitem{bau2020understanding}
David Bau, Jun-Yan Zhu, Hendrik Strobelt, Agata Lapedriza, Bolei Zhou, and
  Antonio Torralba.
\newblock Understanding the role of individual units in a deep neural network.
\newblock {\em PNAS}, 2020.

\bibitem{bau2019gandissect}
David Bau, Jun-Yan Zhu, Hendrik Strobelt, Bolei Zhou, Joshua~B. Tenenbaum,
  William~T. Freeman, and Antonio Torralba.
\newblock Gan dissection: Visualizing and understanding generative adversarial
  networks.
\newblock In {\em ICLR}, 2019.

\bibitem{brock2018large}
Andrew Brock, Jeff Donahue, and Karen Simonyan.
\newblock Large scale {GAN} training for high fidelity natural image synthesis.
\newblock In {\em ICLR}, 2019.

\bibitem{Carreira_2017_CVPR}
Joao Carreira and Andrew Zisserman.
\newblock Quo vadis, action recognition? a new model and the kinetics dataset.
\newblock In {\em CVPR}, 2017.

\bibitem{chan2019everybody}
Caroline Chan, Shiry Ginosar, Tinghui Zhou, and Alexei~A Efros.
\newblock Everybody dance now.
\newblock In {\em ICCV}, 2019.

\bibitem{cho2014learning}
Kyunghyun Cho, Bart Van~Merri{\"e}nboer, Caglar Gulcehre, Dzmitry Bahdanau,
  Fethi Bougares, Holger Schwenk, and Yoshua Bengio.
\newblock Learning phrase representations using rnn encoder-decoder for
  statistical machine translation.
\newblock {\em EMNLP}, 2014.

\bibitem{cohen2019gauge}
Taco~S Cohen, Maurice Weiler, Berkay Kicanaoglu, and Max Welling.
\newblock Gauge equivariant convolutional networks and the icosahedral cnn.
\newblock {\em arXiv preprint arXiv:1902.04615}, 2019.

\bibitem{ebert2017self}
Frederik Ebert, Chelsea Finn, Alex~X Lee, and Sergey Levine.
\newblock Self-supervised visual planning with temporal skip connections.
\newblock {\em arXiv preprint arXiv:1710.05268}, 2017.

\bibitem{feichtenhofer2019slowfast}
Christoph Feichtenhofer, Haoqi Fan, Jitendra Malik, and Kaiming He.
\newblock Slowfast networks for video recognition.
\newblock In {\em CVPR}, 2019.

\bibitem{goetschalckx2019ganalyze}
Lore Goetschalckx, Alex Andonian, Aude Oliva, and Phillip Isola.
\newblock Ganalyze: Toward visual definitions of cognitive image properties.
\newblock In {\em ICCV}, pages 5744--5753, 2019.

\bibitem{goodfellow2014generative}
Ian Goodfellow, Jean Pouget-Abadie, Mehdi Mirza, Bing Xu, David Warde-Farley,
  Sherjil Ozair, Aaron Courville, and Yoshua Bengio.
\newblock Generative adversarial nets.
\newblock In {\em NIPS}, 2014.

\bibitem{hara2018can}
Kensho Hara, Hirokatsu Kataoka, and Yutaka Satoh.
\newblock {Can Spatiotemporal 3D CNNs Retrace the History of 2D CNNs and
  ImageNet?}
\newblock In {\em CVPR}, 2018.

\bibitem{NIPS2017_7240}
Martin Heusel, Hubert Ramsauer, Thomas Unterthiner, Bernhard Nessler, and Sepp
  Hochreiter.
\newblock Gans trained by a two time-scale update rule converge to a local nash
  equilibrium.
\newblock In {\em NIPS}, 2017.

\bibitem{hinton2011transforming}
Geoffrey~E Hinton, Alex Krizhevsky, and Sida~D Wang.
\newblock Transforming auto-encoders.
\newblock In {\em ICANN}. Springer, 2011.

\bibitem{isola2017image}
Phillip Isola, Jun-Yan Zhu, Tinghui Zhou, and Alexei~A Efros.
\newblock {Image-to-Image Translation with Conditional Adversarial Networks}.
\newblock In {\em CVPR}, 2017.

\bibitem{gansteerability}
Ali Jahanian, Lucy Chai, and Phillip Isola.
\newblock On the "steerability" of generative adversarial networks.
\newblock In {\em ICLR}, 2020.

\bibitem{jang2018video}
Yunseok Jang, Gunhee Kim, and Yale Song.
\newblock {Video Prediction with Appearance and Motion Conditions}.
\newblock In {\em ICML}, 2018.

\bibitem{karras2017progressive}
Tero Karras, Timo Aila, Samuli Laine, and Jaakko Lehtinen.
\newblock Progressive growing of gans for improved quality, stability, and
  variation.
\newblock {\em arXiv preprint arXiv:1710.10196}, 2017.

\bibitem{karras2019style}
Tero Karras, Samuli Laine, and Timo Aila.
\newblock A style-based generator architecture for generative adversarial
  networks.
\newblock In {\em CVPR}, 2019.

\bibitem{Karras2019stylegan2}
Tero Karras, Samuli Laine, Miika Aittala, Janne Hellsten, Jaakko Lehtinen, and
  Timo Aila.
\newblock Analyzing and improving the image quality of {StyleGAN}.
\newblock In {\em CVPR}, 2020.

\bibitem{kingma2014adam}
Diederik~P Kingma and Jimmy Ba.
\newblock Adam: A method for stochastic optimization.
\newblock {\em arXiv preprint arXiv:1412.6980}, 2014.

\bibitem{ledig2017photo}
Christian Ledig, Lucas Theis, Ferenc Husz{\'a}r, Jose Caballero, Andrew
  Cunningham, Alejandro Acosta, Andrew~P Aitken, Alykhan Tejani, Johannes Totz,
  Zehan Wang, et~al.
\newblock Photo-realistic single image super-resolution using a generative
  adversarial network.
\newblock In {\em CVPR}, 2017.

\bibitem{lenc2015understanding}
Karel Lenc and Andrea Vedaldi.
\newblock Understanding image representations by measuring their equivariance
  and equivalence.
\newblock In {\em CVPR}, 2015.

\bibitem{li2018flow}
Yijun Li, Chen Fang, Jimei Yang, Zhaowen Wang, Xin Lu, and Ming-Hsuan Yang.
\newblock Flow-grounded spatial-temporal video prediction from still images.
\newblock In {\em ECCV}, 2018.

\bibitem{ma2018disentangled}
Liqian Ma, Qianru Sun, Stamatios Georgoulis, Luc Van~Gool, Bernt Schiele, and
  Mario Fritz.
\newblock Disentangled person image generation.
\newblock In {\em CVPR}, 2018.

\bibitem{Mescheder2018ICML}
Lars Mescheder, Sebastian Nowozin, and Andreas Geiger.
\newblock Which training methods for gans do actually converge?
\newblock In {\em ICML}, 2018.

\bibitem{miyato2018spectral}
Takeru Miyato, Toshiki Kataoka, Masanori Koyama, and Yuichi Yoshida.
\newblock Spectral normalization for generative adversarial networks.
\newblock In {\em ICLR}, 2018.

\bibitem{Nagrani19}
Arsha Nagrani, Joon~Son Chung, Weidi Xie, and Andrew Zisserman.
\newblock Voxceleb: Large-scale speaker verification in the wild.
\newblock {\em Computer Science and Language}, 2019.

\bibitem{ohnishi2018ftgan}
Katsunori Ohnishi, Shohei Yamamoto, Yoshitaka Ushiku, and Tatsuya Harada.
\newblock Hierarchical video generation from orthogonal information: Optical
  flow and texture.
\newblock In {\em AAAI}, 2018.

\bibitem{pan2019video}
Junting Pan, Chengyu Wang, Xu Jia, Jing Shao, Lu Sheng, Junjie Yan, and
  Xiaogang Wang.
\newblock Video generation from single semantic label map.
\newblock {\em arXiv preprint arXiv:1903.04480}, 2019.

\bibitem{park2020cut}
Taesung Park, Alexei~A. Efros, Richard Zhang, and Jun-Yan Zhu.
\newblock Contrastive learning for unpaired image-to-image translation.
\newblock In {\em ECCV}, 2020.

\bibitem{park2019semantic}
Taesung Park, Ming-Yu Liu, Ting-Chun Wang, and Jun-Yan Zhu.
\newblock Semantic image synthesis with spatially-adaptive normalization.
\newblock In {\em Proceedings of the IEEE Conference on Computer Vision and
  Pattern Recognition}, pages 2337--2346, 2019.

\bibitem{paszke2019pytorch}
Adam Paszke, Sam Gross, Francisco Massa, Adam Lerer, James Bradbury, Gregory
  Chanan, Trevor Killeen, Zeming Lin, Natalia Gimelshein, Luca Antiga, et~al.
\newblock Pytorch: An imperative style, high-performance deep learning library.
\newblock In {\em NeurIPS}, 2019.

\bibitem{saito2017temporal}
Masaki Saito, Eiichi Matsumoto, and Shunta Saito.
\newblock Temporal generative adversarial nets with singular value clipping.
\newblock In {\em ICCV}, 2017.

\bibitem{TGAN2020}
Masaki Saito, Shunta Saito, Masanori Koyama, and Sosuke Kobayashi.
\newblock Train sparsely, generate densely: Memory-efficient unsupervised
  training of high-resolution temporal gan.
\newblock {\em IJCV}, 2020.

\bibitem{shaham2019singan}
Tamar~Rott Shaham, Tali Dekel, and Tomer Michaeli.
\newblock Singan: Learning a generative model from a single natural image.
\newblock In {\em CVPR}, 2019.

\bibitem{shen2020interpreting}
Yujun Shen, Jinjin Gu, Xiaoou Tang, and Bolei Zhou.
\newblock Interpreting the latent space of gans for semantic face editing.
\newblock In {\em CVPR}, 2020.

\bibitem{soomro2012ucf101}
Khurram Soomro, Amir~Roshan Zamir, and Mubarak Shah.
\newblock {UCF101: A Dataset of 101 Human Action Classes From Videos in The
  Wild}.
\newblock Technical report, CRCV-TR-12-01, November 2012.

\bibitem{tran2015learning}
Du Tran, Lubomir Bourdev, Rob Fergus, Lorenzo Torresani, and Manohar Paluri.
\newblock Learning spatiotemporal features with 3d convolutional networks.
\newblock In {\em ICCV}, 2015.

\bibitem{tulyakov2017mocogan}
Sergey Tulyakov, Ming-Yu Liu, Xiaodong Yang, and Jan Kautz.
\newblock {MoCoGAN}: Decomposing motion and content for video generation.
\newblock In {\em CVPR}, 2018.

\bibitem{varol19_surreact}
G{\"u}l Varol, Ivan Laptev, Cordelia Schmid, and Andrew Zisserman.
\newblock Synthetic humans for action recognition from unseen viewpoints.
\newblock {\em CoRR}, abs/1912.04070, 2019.

\bibitem{varol17_surreal}
G{\"u}l Varol, Javier Romero, Xavier Martin, Naureen Mahmood, Michael~J. Black,
  Ivan Laptev, and Cordelia Schmid.
\newblock Learning from synthetic humans.
\newblock In {\em CVPR}, 2017.

\bibitem{vaswani2017attention}
Ashish Vaswani, Noam Shazeer, Niki Parmar, Jakob Uszkoreit, Llion Jones,
  Aidan~N Gomez, {\L}ukasz Kaiser, and Illia Polosukhin.
\newblock Attention is all you need.
\newblock In {\em NeurIPS}, 2017.

\bibitem{vondrick2016generating}
Carl Vondrick, Hamed Pirsiavash, and Antonio Torralba.
\newblock Generating videos with scene dynamics.
\newblock In {\em NIPS}, 2016.

\bibitem{voynov2020unsupervised}
Andrey Voynov and Artem Babenko.
\newblock Unsupervised discovery of interpretable directions in the gan latent
  space.
\newblock {\em arXiv preprint arXiv:2002.03754}, 2020.

\bibitem{walker2017pose}
Jacob Walker, Kenneth Marino, Abhinav Gupta, and Martial Hebert.
\newblock The pose knows: Video forecasting by generating pose futures.
\newblock In {\em ICCV}, 2017.

\bibitem{wang2019fewshotvid2vid}
Ting-Chun Wang, Ming-Yu Liu, Andrew Tao, Guilin Liu, Jan Kautz, and Bryan
  Catanzaro.
\newblock Few-shot video-to-video synthesis.
\newblock In {\em NeurIPS}, 2019.

\bibitem{wang2018vid2vid}
Ting-Chun Wang, Ming-Yu Liu, Jun-Yan Zhu, Guilin Liu, Andrew Tao, Jan Kautz,
  and Bryan Catanzaro.
\newblock Video-to-video synthesis.
\newblock In {\em NeurIPS}, 2018.

\bibitem{wang2018high}
Ting-Chun Wang, Ming-Yu Liu, Jun-Yan Zhu, Andrew Tao, Jan Kautz, and Bryan
  Catanzaro.
\newblock High-resolution image synthesis and semantic manipulation with
  conditional gans.
\newblock In {\em Proceedings of the IEEE conference on computer vision and
  pattern recognition}, pages 8798--8807, 2018.

\bibitem{wang2020g3an}
Yaohui Wang, Piotr Bilinski, Francois Bremond, and Antitza Dantcheva.
\newblock {G3AN}: Disentangling appearance and motion for video generation.
\newblock In {\em CVPR}, 2020.

\bibitem{Wang_2020_CVPR}
Yaohui Wang, Piotr Bilinski, Francois Bremond, and Antitza Dantcheva.
\newblock {G3AN}: Disentangling appearance and motion for video generation.
\newblock In {\em Proceedings of the IEEE/CVF Conference on Computer Vision and
  Pattern Recognition (CVPR)}, June 2020.

\bibitem{xu2018attngan}
Tao Xu, Pengchuan Zhang, Qiuyuan Huang, Han Zhang, Zhe Gan, Xiaolei Huang, and
  Xiaodong He.
\newblock Attngan: Fine-grained text to image generation with attentional
  generative adversarial networks.
\newblock In {\em CVPR}, 2018.

\bibitem{yang2018pose}
Ceyuan Yang, Zhe Wang, Xinge Zhu, Chen Huang, Jianping Shi, and Dahua Lin.
\newblock Pose guided human video generation.
\newblock In {\em ECCV}, 2018.

\bibitem{yang2020temporal}
Ceyuan Yang, Yinghao Xu, Jianping Shi, Bo Dai, and Bolei Zhou.
\newblock Temporal pyramid network for action recognition.
\newblock In {\em CVPR}, 2020.

\bibitem{yu2018bisenet}
Changqian Yu, Jingbo Wang, Chao Peng, Changxin Gao, Gang Yu, and Nong Sang.
\newblock Bisenet: Bilateral segmentation network for real-time semantic
  segmentation.
\newblock In {\em Proceedings of the European conference on computer vision
  (ECCV)}, 2018.

\bibitem{zakharov2019few}
Egor Zakharov, Aliaksandra Shysheya, Egor Burkov, and Victor Lempitsky.
\newblock Few-shot adversarial learning of realistic neural talking head
  models.
\newblock In {\em ICCV}, 2019.

\bibitem{zhao2019image}
Bo Zhao, Lili Meng, Weidong Yin, and Leonid Sigal.
\newblock Image generation from layout.
\newblock In {\em CVPR}, 2019.

\bibitem{Zhao_2018_ECCV}
Long Zhao, Xi Peng, Yu Tian, Mubbasir Kapadia, and Dimitris Metaxas.
\newblock Learning to forecast and refine residual motion for image-to-video
  generation.
\newblock In {\em ECCV}, 2018.

\bibitem{CycleGAN2017}
Jun-Yan Zhu, Taesung Park, Phillip Isola, and Alexei~A Efros.
\newblock Unpaired image-to-image translation using cycle-consistent
  adversarial networkss.
\newblock In {\em ICCV}, 2017.

\end{thebibliography}
